\documentclass[sigconf,nonacm]{acmart}

\usepackage{CJKutf8}

\usepackage{array}

\usepackage[utf8]{inputenc} 

\usepackage[T1]{fontenc}    

\usepackage{hyperref}       

\usepackage{url}            

\usepackage{booktabs}       

\usepackage{amsfonts}       

\usepackage{nicefrac}       

\usepackage{microtype}      

\usepackage[table]{xcolor}  

\usepackage{graphicx}

\usepackage{amsmath}

\usepackage{wrapfig}

\usepackage{color}

\usepackage{multirow}

\usepackage{enumitem}

\usepackage{placeins}

\usepackage[ruled,vlined]{algorithm2e}

\usepackage{comment}

\usepackage{caption}

\usepackage{subcaption}

\usepackage{pifont}

\newcommand{\ourmethod}{\textit{SwiftExplorer}}

\newcommand{\free}{training-free}

\newcommand{\dive}{diversity}

\newcommand{\fid}{fidelity}

\usepackage[normalem]{ulem}

\definecolor{myyellow}{RGB}{190,144,0}

\definecolor{mygreen}{RGB}{0,136,51}

\definecolor{myblue}{RGB}{0,102,204}

\begin{document}

\begin{CJK*}{UTF8}{gbsn}


\title{\ourmethod{}: Training-free Diffusion Model Alignment with Swift Diversity Exploration}


\author{Renye Yan}
\authornote{Renye Yan and Jikang Cheng contributed equally to this work.}
\affiliation{%
  \institution{Peking University}
  \city{Beijing}
  \country{China}
}

\author{Jikang Cheng}
\authornotemark[1]
\affiliation{%
  \institution{Peking University}
  \city{Beijing}
  \country{China}
}

\author{You Wu}
\affiliation{%
  \institution{Nanjing University}
  \city{Nanjing}
  \country{China}
}

\author{Baojin Huang}
\affiliation{%
  \institution{Huazhong Agricultural University}
  \city{Wuhan}
  \country{China}
}

\author{Wei Peng}
\affiliation{%
  \institution{Stanford University}
  \city{Stanford}
  \state{California}
  \country{USA}
}

\author{Zongwei Wang}
\affiliation{%
  \institution{Peking University}
  \city{Beijing}
  \country{China}
}


\author{Ling Liang}
\authornote{Ling Liang and Yimao Cai are the corresponding authors.}
\affiliation{%
  \institution{Peking University}
  \city{Beijing}
  \country{China}
}

\author{Yimao Cai}
\authornotemark[2]
\affiliation{%
  \institution{Peking University}
  \city{Beijing}
  \country{China}
}


\renewcommand{\shortauthors}{Yan et al.}

\begin{abstract}

Diffusion models have general generative abilities but struggle to align with specific objectives. Fine-tuning can improve alignment, yet its training cost is often prohibitive. This led to \free{} methods that apply objective-guided terms in sampling to bias the generation distribution toward designated regions, e.g., high-reward areas. However, these methods face two issues: (1) the strong directional bias narrows the pretrained distribution and generation diversity, and (2) indiscriminate constant guidance fails to prune redundant signals, hurting both quality and efficiency. To address the above challenges, we propose \ourmethod{}, a plugin that mitigates distribution collapse caused by excessive diversity loss and reduces compute costs. First, we adopt an Inheritance-Restart exploration mechanism to avoid early convergence, while exploration also increases the likelihood of high-reward trajectories. Additionally, it balances \dive{} and \fid{}, adding diversity without causing a distribution over-shift. Second, our Quality--Efficiency arbitration mechanism improves guidance by removing incorrect signals, and it reduces computation by dynamically stopping generation when completeness and marginal reward gain are optimal.
In an extensive number of experiments and different types of evaluation metrics, the proposed \ourmethod{} achieves excellent performance on all metrics, including preference, fidelity, diversity, and richness.

\end{abstract}







\maketitle

\section{INTRODUCTION}
\begin{figure}
    \centering
    \includegraphics[width=1\linewidth]{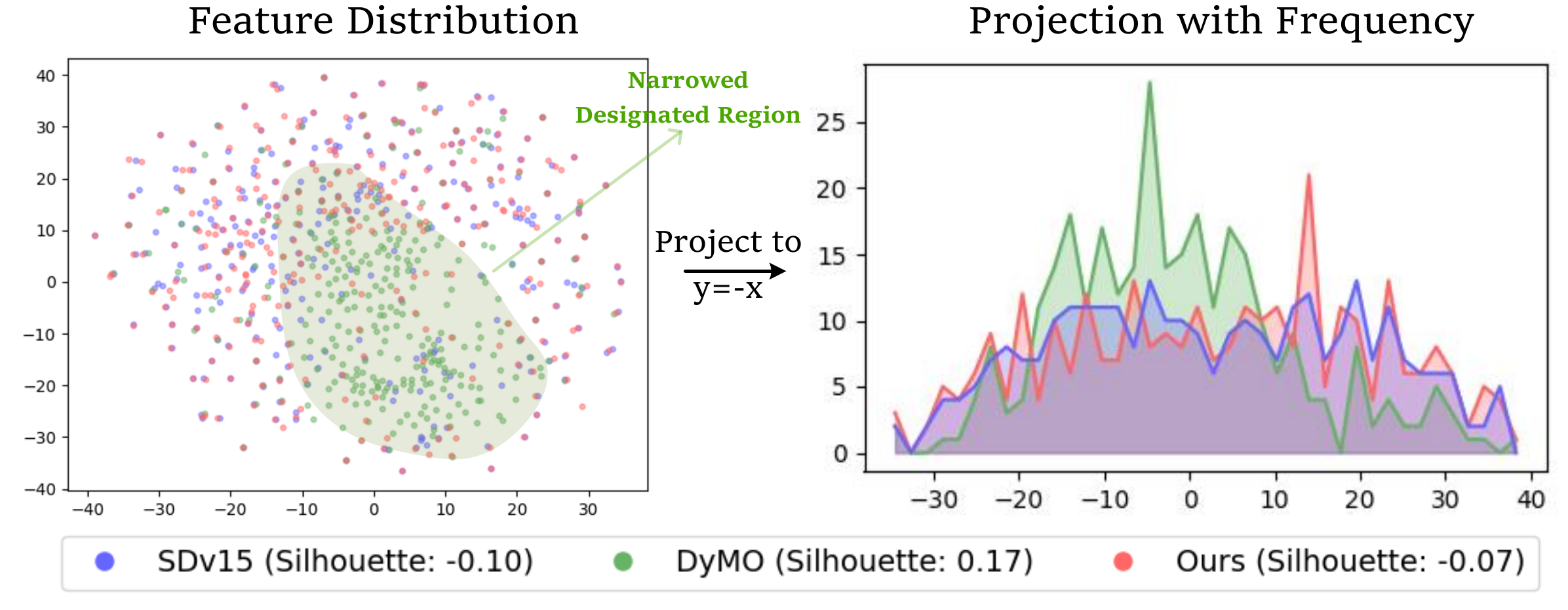}
    \caption{\textbf{Feature visualization via t-SNE}. The features are extracted by a general self-supervised decoder (\textit{i.e.}, DINOv2) from images generated by different methods. With the guidance targets of both Ours and DyMO having achieved comparable scores, the distribution results indicate that existing SOTA guidance methods (\textit{i.e.}, DyMO) yield more compact clustering, narrowing the generated distribution, which in turn reduces the diversity of the generated data. This observation is substantiated by the Silhouette Coefficient~\cite{silh}, a clustering evaluation metric, which demonstrates its convergence and limited diversity. In contrast, our method maintains a broader generative distribution while achieving comparable target reward scores (see Tab.~\ref{tab:main}).}
    \label{fig:instr}
    \vspace{-0.3cm}
\end{figure}

Diffusion models~\cite{ho2020denoising,ramesh2022hierarchical,rombach2022ldm,zhang2026rep,lu2026chordedit,yan2025entropy,yan2026pixel,yan2026less,li2026videococo,ling2026gentrack,ma2026can,shen2025generative}, as one of the most powerful generative frameworks, demonstrate exceptional generative capabilities by training on various large-scale datasets. This is evident in the model's capability to generate both high-quality images and a diverse range of semantic and stylistic content. However, in practical applications, it is not only necessary to generate realistic images but also to ensure that the results align with specific target requirements, such as aesthetic preferences. This need for goal alignment is becoming a core research problem, posing the challenge of effectively translating the general capabilities of diffusion models into customized generation.


Early research primarily relied on training strategies~\cite{black2023training,fan2024reinforcement,yang2024using,wallace2024diffusion}. By fine-tuning diffusion models on specific preference objectives, alignment performance could be significantly improved. However, such methods generally face challenges of high training costs and poor transferability~\cite{kim2025test,xie2025dymo}. 
Whenever user requirements or objectives change, diffusion models must be finetuned again, which limits their scalability and transferability in practical applications.
To address this, researchers proposed training-free methods~\cite{chung2022diffusion,kim2025test,deckers2024manipulating,tang2024inference,xie2025dymo}, which do not modify the model's weight but incorporate external signals during generation to achieve goal-guided results. The advantages of such approaches are evident as they require no additional training costs and can quickly adapt to different objective demands. Consequently, training-free guidance has become a crucial diffusion model alignment method. More detailed \emph{\textbf{Related Works}} can be found in Appendix~\ref{related work}.


However, existing training-free methods still suffer from two typical limitations: (1) \underline{\textit{overly narrow generative distributions}}. To enhance the alignment between generated results and target objectives, training-free methods should incorporate target-guided terms during sampling. These methods gradually shift the generative distribution toward specific regions, such as high-reward subspaces. Although these approaches effectively enhance alignment, their excessive reliance on a consistent guidance direction easily leads to reward hacking~\cite{tang2024inference}, which manifests as a rapid narrowing of the pretrained distribution, ultimately reducing image diversity~\cite{ho2022classifier,tang2024inference}. (2) \underline{\textit{unselective, inefficient guidance}}. Current methods apply guidance terms indiscriminately throughout the entire generation process, lacking the ability to identify and manage ineffective or redundant signals. This not only results in unnecessary computational waste but may also introduce noise interference due to ineffective guidance, ultimately compromising both alignment quality and generation efficiency.

To address the above challenges, we propose \ourmethod{}, an enhanced plugin compatible with existing training-free alignment methods. (1) \ourmethod{} introduces an Inheritance-Restart exploration mechanism that prevents premature convergence of generation trajectories to a single mode through probabilistic perturbation and path screening. Simultaneously, we introduce a progress-aware adaptive trade-off to enhance exploration while preserving generation fidelity (see Fig.~\ref{fig:instr}). It promotes early-stage exploration to improve diversity and the discovery of high-reward trajectories. In later stages, it encourages convergence to preserve the pretrained distribution. (2) To improve guidance efficiency, \ourmethod{} employs a Quality-Efficiency screening mechanism to filter out ineffective signals, ensuring that each guidance step contributes meaningfully. Additionally, by integrating an adaptive factor and considering the diminishing marginal utility of rewards, our method dynamically cuts guidance in later generations to avoid redundant optimization. The Quality-Efficiency mechanism effectively reduces computational overhead while maintaining alignment quality. Experiments show that while ensuring leading performance on both fidelity and alignment objectives, \ourmethod{} effectively improves generative diversity and outperforms existing baselines in computational efficiency. Our contributions are as follows: 

\begin{itemize}[leftmargin=2em, labelsep=0.5em]
    \item We propose an Inheritance-Restart exploration mechanism to enhance the generation diversity of existing  methods.
    \item We introduce a Quality-Efficiency mechanism to drop invalid signals, thereby improving guidance and reducing costs.
    \item We achieve two trade-offs: Diversity and Fidelity evaluation during generation progress. Quality and Efficiency adjustment on marginal reward gains.
\end{itemize}

\section{Method}
\begin{figure*}[t!]
    \centering
    \includegraphics[width=1\linewidth]{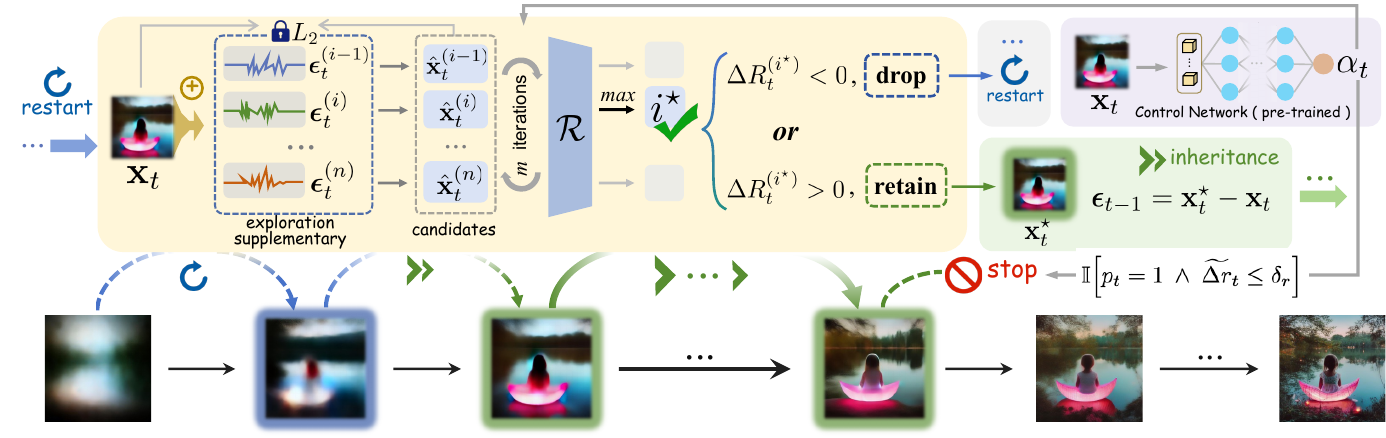}
    \caption{\textbf{Overall framework} of the proposed plug-and-play \ourmethod{}.}
    \label{fig:framework}
\end{figure*}



\ourmethod{} is a plugin enhancing training-free methods in exploration and efficiency.
Sec.~\ref{per} covers preliminaries.
Sec.~\ref{inherit} elaborates on the Inheritance-Restart exploration mechanism, achieving an adaptive trade-off between Diversity and Fidelity.
Sec.~\ref{quality} introduces a Quality–Efficiency arbitration mechanism guiding early quality and late efficiency.
The \ourmethod{} framework is shown in Fig.~\ref{fig:framework}, with algorithmic details in Appendix.~\ref{sec:alg}.

\subsection{Reward-Guided Diffusion Alignment}
\label{per}
\subsubsection{Generative Process of Diffusion Models}
The discrete-time generative process (or reverse process) aims to produce a sample by denoising a latent variable $\mathbf{x}_t$ iteratively. This process requires two components: a predefined noise schedule $\{\beta_t\}_{t=0}^T$ and a well-trained score function $\mathbf{s}_\theta(\mathbf{x}_t, t)$. We define $\alpha_t = 1 - \beta_t$ and $\bar{\alpha}_t = \prod_{i=1}^t \alpha_i$.
The generation starts by sampling an initial latent variable from a standard normal distribution:
$ \mathbf{x}_T \sim \mathcal{N}(\mathbf{0}, \mathbf{I}) $
Then, for each timestep $t$ from $T$ to $0$, the sample is iteratively refined using the update rule:
\begin{equation}
\label{eq:iterative_generation}
    \mathbf{x}_{t-1} = \frac{1}{\sqrt{\alpha_t}} \left( \mathbf{x}_t + \beta_t \mathbf{s}_\theta(\mathbf{x}_t, t) \right) + \sqrt{\beta_t} \mathbf{z}_t,
\end{equation}
where $\mathbf{z}_t \sim \mathcal{N}(\mathbf{0}, \mathbf{I})$ is a standard Gaussian noise vector independently sampled at each timestep. The final output after $T$ steps is the generated sample $\mathbf{x}_0$. 
For convenience, we also define a function ${f}_\theta(\mathbf{x}_t, t)$ that provides a direct estimate of the clean sample $\mathbf{x}_{0|t}$, from any noisy intermediate $\mathbf{x}_t$~\cite{chung2022diffusion,bansal2023universal,xie2025dymo}:
\begin{equation}
\label{X0t}
    \mathbf{x}_{0|t} = {f}_\theta(\mathbf{x}_t, t) = \frac{1}{\sqrt{\bar{\alpha}_t}}\left(\mathbf{x}_t + (1 - \bar{\alpha}_t)\mathbf{s}_\theta(\mathbf{x}_t, t)\right) .
\end{equation}


\subsubsection{Iterative training-free Refinement with Gradient Guidance}
To guide the generation process towards a specific objective, we can incorporate the gradient of a reward function $\mathcal{R}$~\cite{li2025preference}. The reward function is defined in the data domain (i.e., it operates on the estimated clean sample ${\mathbf{x}}_{0|t}$ by Eq.~\ref{X0t}. The approach of predicting ${\mathbf{x}}_{0|t}$ using Eq.~\ref{X0t} for reward calculation is widely utilized in the field~\cite{xie2025dymo,bansal2023universal}).

The guidance is injected by adjusting the current sample $\mathbf{x}_t$ in the direction of the reward gradient. Specifically, before applying the update in Eq.~\ref{eq:iterative_generation}, we compute a guided sample ${\mathbf{x}}_t$:
\begin{equation}
   {\mathbf{x}}_t \leftarrow \mathbf{x}_t + w \nabla_{\mathbf{x}_t}\mathcal{R}({f}_\theta(\mathbf{x}_t, t)), \label{eq:one-step}
\end{equation}

where $w$ is a guidance scale factor, $\mathcal{R}(\cdot)$ denotes the reward function for inference-time guidance.
{The above process could be seen as one step of optimization. For simplicity, we can define optimization with $m$ steps of Eq.~\ref{eq:one-step} as:
\begin{equation}
   \tilde{\mathbf{x}}_t = g(\mathbf{x}_t,\mathcal{R},m). \label{eq:iter}
\end{equation}
}
Replacing $\mathbf{x}_t$ in Eq.~\ref{eq:iterative_generation} with 
\( \tilde{\mathbf{x}}_t \), the new update rule becomes:
\begin{equation}
\label{eq:guided_generation}
    \mathbf{x}_{t-1} = \frac{1}{\sqrt{\alpha_t}} \left( {\tilde{\mathbf{x}}}_t + \beta_t \mathbf{s}_\theta(\tilde{\mathbf{x}}_t, t) \right) + \sqrt{\beta_t} \mathbf{z}_t.
\end{equation}

\subsection{Inheritance–Restart Exploration mechanism}
\label{inherit}

To mitigate the issue of narrowed generative distribution caused by overly concentrated guidance direction in training-free methods, \ourmethod{} equips an Inheritance-Restart exploration mechanism. This mechanism selectively introduces exploratory supplementary terms based on the generation progress, thereby preventing premature convergence of generative trajectories to a single mode and increasing the probability of discovering high-reward paths.


\subsubsection{Exploration Supplement Variable} 
\label{exp}

During the generation step $t$, we introduce an exploration supplement variable $\boldsymbol{\epsilon}^{(i)}_t$, which allows the generated trajectories to maintain a richer distribution. Specifically, we adopt Monte Carlo sampling to obtain $n$ candidates $\quad \boldsymbol{\epsilon}^{(i)}_t \sim \mathcal{N}(0, \mathbf{\sigma}_t^2 I), \quad i=1,\dots,n$. After incorporating $\boldsymbol{\epsilon}^{(i)}_t$, the original 
$\textbf{x}_t$ without exploration is updated into a set $\{\hat{\textbf{x}}_t^{(i)}\}$:
\begin{equation}
  \hat{\mathbf{x}}_t^{(i)} = \mathbf{x}_t + \boldsymbol{\epsilon}^{(i)}_t.   \label{eq:add}
\end{equation}
Subsequently, each candidate $\hat{\mathbf{x}}_t^{(i)}$ is mapped to its predicted reconstruction \(\hat{\mathbf{x}}_{0|t}^{(i)} = f_\theta(\hat{\mathbf{x}}_t^{(i)}, t)\), and is evaluated via the objective reward function \(\mathcal{R}(\hat{\mathbf{x}}_{0|t}^{(i)})\). After $m$ rounds 
based on Eq.~\ref{eq:iter}, we obtain a final set of candidates $\{\tilde{\mathbf{x}}_{t}^{(i)}\}=g(\{\hat{\mathbf{x}_t}^{(i)}\},\mathcal{R},m)$. We then define the optimal generation trajectory policy as:
\begin{equation}
i^\star = \arg\max_{i \in \{1,\dots,n\}} \mathcal{R}(\tilde{\mathbf{x}}_{0|t}^{(i)}), \quad \mathbf{x}_t^\star = \tilde{\mathbf{x}}_t^{(i^\star)}. \label{eq:select}
\end{equation}
This process selects the optimal candidate \(\mathbf{x}_t^\star\) as the sample of generation step $t$, which is adopted for the following generations.
In summary, the injection of the exploration supplementary term expands the local search space. The selection policy picks the optimal generation trajectory with the highest alignment potential, thereby increasing diversity while maximizing alignment rewards.

\subsubsection{Inheritance–Restart Technique}



In Sec.~\ref{exp}, the exploration supplementary term expands the generation space. Building upon this, we apply an Inheritance-Restart technique to this term. This technique dynamically regulates exploration behavior based on the target reward improvement, preserving high-quality explorations while eliminating ineffective adjustments. Consequently, the exploration term enhances diversity, while the Inheritance-Restart technique ensures that exploration consistently improves the target reward. Specifically, the reward changing for the supplementary term at generation step $t$ is defined as: 
\begin{equation}
\Delta \mathcal{R}_t^{(i^{\star})} = \mathcal{R}({\mathbf{x}}_{0|t}^{\star}) - \mathcal{R}({\mathbf{x}}_{0|t}).
\end{equation}

\begin{description}[leftmargin=2em, labelsep=0.5em]

    \item[(1)] If $\Delta \mathcal{R}_t^{(i^{\star})}>0$, 
    the exploration still yields a positive gain for the alignment reward. In this case, the latent change from the previous step is carried over to the next step $t-1$, where $\boldsymbol{\epsilon}_{t-1} = \mathbf{x}_t^{\star} - {\mathbf{x}}_t$. Under this inheritance strategy, the exploration candidate set contains only this inherited signal, i.e., setting $n=1$ in Eq.~\ref{eq:add}, and optimization continues via Eq.~\ref{eq:iter}.
    

    \item[(2)] If $\Delta \mathcal{R}_t^{(i^{\star})}<0$, 
    the exploration gain becomes saturated or starts to decline. In this case, the restart mechanism is activated: $n$ exploration candidates are resampled in the next step $t-1$ as Eq.~\ref{eq:add}, and the exploration term is updated through the optimization in Eq.~\ref{eq:iter} and the selection step in Eq.~\ref{eq:select}.
\end{description}

The inheritance-restart technique uses reward changes as a criterion to enhance exploration diversity and reduce guidance overhead. Moreover, it ensures the generation alignment with target objective.


\subsubsection{Diversity and Fidelity Adaptive Trade-off}

Diffusion models face the dual demands of diversity and fidelity. Diversity expands the coverage of potential generation distribution, while fidelity ensures the results are credible and semantically consistent with the conditioning. However, these two objectives often exhibit a trade-off,
making their careful balance crucial for achieving high-quality results ~\cite{dhariwal2021beat}. To address this, we propose an adaptive coordination technique that dynamically adjusts the emphasis based on the generation progress: it prioritizes exploration in the early stages to foster diversity and gradually increases the strength of regularization in the later stages to ensure fidelity. Unlike approaches that use predefined static parameters, the adaptation technique can flexibly accommodate the dynamic nature of the generation process. 


To this end, we construct a Control Network (This network outputs a simple noise scorer, analogous to scoring models such as PickScore and Aesthetic Score, and is therefore used as a \textbf{\textit{pretrained model}}. Experimental support is provided in Tab.~\ref{tab:control}, with details in Appendix~\ref{control_network}.),
which is designed to assess the denoising progress of the intermediate state in real-time during the generation process. Specifically, the control network is defined as a mapping function:
$h_{\theta_p}: \mathbf{x}_t \mapsto p_t$, where $\mathbf{x}_t \in \mathbb{R}^d$ denotes the latent state at generation step $t$. $p_t \in [0,1]$ represents the probability of the current latent state close to the clean image $\mathbf{x}_0$, which reflects the degree of the current state's generation progress. The control network is obtained by minimizing a binary cross-entropy loss as the supervision signal (see Appendix~\ref{pernet} for details). We utilize $p_t$ as an adaptive weighting to balance exploration diversity and generation fidelity.
In the first optimization round, the combined effect of the exploration and alignment signal gradients is written as: 
\begin{equation}
\hat{\mathbf{x}}_t^{(i)} = {\mathbf{x}}_t + \underbrace{(1-p_t) \cdot \boldsymbol{\epsilon}^{(i)} - p_t\cdot \nabla_{\mathbf{x}_t} L_2(\hat{\mathbf{x}}_t,{\mathbf{x}}_t)}_{\text{Diversity-Fidelity Trade-off}} + \nabla_{\mathbf{x}_t} \mathcal{R}({\mathbf{x}}_{0|t}),
\end{equation}

\begin{figure*}[t!]
    \centering
    \includegraphics[width=0.95\linewidth]{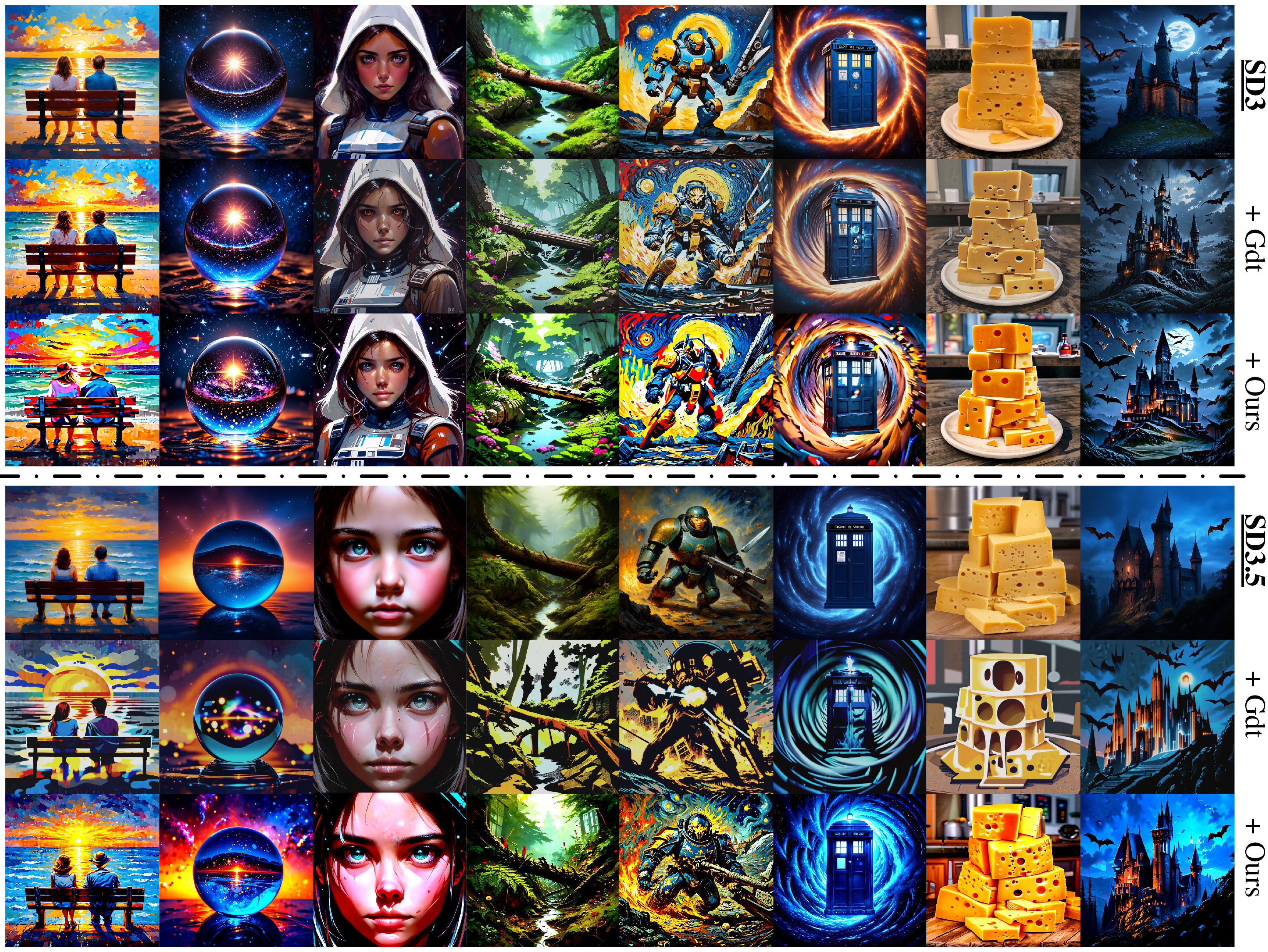}
    \caption{Visual impression on the \textbf{advanced flow-matching-based models}. Our results exhibit finer details, more diverse colors, and improved aesthetic style.}
    \label{fig:sd3}
\end{figure*}
where \(\boldsymbol{\epsilon}^{(i)}\) is the exploration supplement variable and \(\nabla_{\mathbf{x}_t}\mathcal{R}({\mathbf{x}}_{0|t})\) represents the reward-based gradient guidance. We introduce L2 regularization to constrain the exploration scope. Then, based on the iteration defined in Eq.~\ref{eq:iter}, we iteratively introduce guidance to obtain the final state $\tilde{\mathbf{x}}_t^{(i)}=g(\hat{\mathbf{x}}_t^{(i)},\mathcal{R}-L_2,m)$. 


In summary, the adaptive trade-off encourages exploratory diversity in the early generation stages. In the later stages, the continuous increase in the regularization constraint prompts the generation process to transition from exploration to stable convergence, thereby ensuring high-fidelity results. Through this design, \ourmethod{} achieves an adaptive trade-off between creativity and stability.

\subsection{Quality–Efficiency Arbitration mechanism}
\label{quality}

\begin{table*}[t]
    \setlength{\tabcolsep}{4pt}
    \centering
    \caption{\textbf{Main comparisons with SoTA}. All metrics are obtained with SDv15 as backbone, Pick-a-pic as prompt set, and PickScore as reward for guidance. \textit{Orange denotes gradient-based training methods. Green denotes reinforcement-learning-based training methods. Blue denotes training-free methods}.}\label{tab:main}
\begin{tabular}{l|c|ccccccccccccc}
    \toprule
    \multirow{2}{*}{Method} & \multicolumn{4}{c}{\textbf{Preference}} & \multicolumn{3}{c}{\textbf{Fidelity}} & \multicolumn{3}{c}{\textbf{Diversity}} & \multicolumn{3}{c}{\textbf{Richness}} & \multirow{2}{*}{\#Top2} \\
    \cmidrule(lr){2-5}\cmidrule(lr){6-8}\cmidrule(lr){9-11}\cmidrule(lr){12-14}
        & PS$\uparrow$ & AES$\uparrow$ & IR$\uparrow$ & HPS$\uparrow$
        & FID$\downarrow$ & CLIP$\uparrow$ & iFS$\uparrow$
        & LPIPS$\uparrow$ & IS$\uparrow$ & TCE$\uparrow$
        & BRI$\downarrow$ & NIQE$\downarrow$ & SE$\uparrow$ & \\
    \midrule
    SDv15 & 20.48 & 5.412 & 0.181 & 0.262 & - & \underline{0.243} & - & \textbf{0.654} & \textbf{23.79} & 38.05 & 18.66 & 5.401 & 11.26 & 3\\
    \midrule
    \rowcolor{orange!10}Diff-DPO & 20.87 & 5.551 & 0.443 & 0.271 & 109.1 & \textbf{0.244} & 0.795 & 0.639 & 22.77 & 39.20 & 15.12 & 4.463 & 11.02 & 1\\
    \rowcolor{orange!10}Diff-KTO & 20.83 & 5.585 & 0.599 & 0.272 & 101.3 & 0.240 & 0.801 & 0.634 & 22.70 & 39.12 & 26.25 & 4.361 & 11.22 & 0\\
    \rowcolor{orange!10}SPO & 20.76 & 5.613 & 0.282 & 0.218 & 78.76 & 0.241 & 0.855 & 0.649 & 23.18 & 39.32 & 25.67 & \underline{4.103} & 11.09 & 1\\
    \rowcolor{orange!10}DRaFT & 22.52 & \underline{5.697} & \underline{0.779} & 0.271 & 84.47 & 0.231 & 0.856 & 0.572 & 20.10 & 37.85 & 35.64 & 5.336 & 10.43 & 2\\
    \midrule
    \rowcolor{green!10}DDPO & 21.79 & 5.704 & 0.196 & 0.212 & 147.5 & 0.242 & 0.539 & 0.629 & 20.01 & 39.18 & 12.46 & 4.328 & \underline{11.74} & 1\\
    \rowcolor{green!10}DPOK & 20.97 & 5.661 & 0.583 & 0.272 & 99.30 & 0.242 & 0.820 & 0.641 & 22.52 & \underline{39.35} & \underline{12.28} & 4.608 & 11.23 & 2\\
    \midrule
    \rowcolor{blue!5}DNO & 20.87 & 5.479 & 0.425 & 0.271 & 79.96 & 0.239 & 0.851 & 0.591 & 19.94 & 38.88 & 17.23 & 4.757 & 11.07 & 0\\
    \rowcolor{blue!5}TTScale & 22.14 & 5.636 & 0.619 & 0.272 & \textbf{66.93} & 0.240 & 0.863 & 0.652 & \underline{23.73} & 38.90 & 15.05 & 4.447 & 11.28 & 2\\
    \rowcolor{blue!5}FKSteer & 22.56 & 5.696 & 0.539 & 0.278 & 73.83 & \underline{0.243} & \underline{0.868} & 0.650 & 23.76 & 39.42 & 13.93 & 4.092 & 11.59 & 2\\
    \rowcolor{blue!5}DAS & 22.34 & 5.658 & 0.632 & \textbf{0.283} & 95.42 & 0.240 & 0.844 & 0.625 & 21.01 & 39.19 & 14.36 & 4.631 & 11.14 & 1\\
    \rowcolor{blue!5}DyMO & \underline{22.79} & 5.694 & 0.709 & 0.279 & 90.56 & 0.239 & 0.849 & 0.627 & 20.97 & 39.29 & 16.18 & 4.545 & 11.29 & 1\\
    \midrule
    \rowcolor{blue!15}\textbf{Ours} & \textbf{22.91} & \textbf{5.707} & \textbf{0.812} & \underline{0.280} & \underline{71.36} & 0.242 & \textbf{0.891} & \textbf{0.654} & \underline{23.71} & \textbf{39.37} & \textbf{10.92} & \textbf{2.886} & \textbf{12.07} & \textcolor{red}{\textbf{12}}\\
    \bottomrule
\end{tabular}
\end{table*}
\subsubsection{Guidance Screening}


Existing methods apply guidance signals indiscriminately throughout the entire sampling process, which can lead to ineffective guidance and undermine alignment effectiveness. To address this, \ourmethod{} introduces a guidance signal selection technique that dynamically discriminates the signal's utility. 
Let the optimal guidance signal generated at step $t$ be \(\boldsymbol{\gamma}_t^{(i^\star)}\), which represents the reward gradient corresponding to $\mathbf{x}_t^\star$, \textit{i.e.}, the optimal candidate. We then define a selection operator \(\mathcal{F}\):
\begin{equation}
    \mathcal{F}(\boldsymbol{\gamma}_t^{(i^\star)}) = \{\Delta \mathcal{R}^{(i^*)}_t >0 \}.
\end{equation}
With the proposed screening process, a guidance signal is retained only if it contributes to a positive reward gain; otherwise, it is discarded, and the original 
$\mathbf{x}_t$ from the base diffusion model is used.
This reward-margin-based selection strategy ensures the quality of the guidance signals for generation. 




\subsubsection{Quality and Compute Adaptive Trade-off}
Unlike existing methods that rigidly couple guidance 
with the generation steps, \ourmethod{} regulates the use of guidance signals through the dual control of target reward marginal gain and generation progress $p_t$. This approach aims to reduce computational costs while ensuring alignment quality. Guidance is stopped when it is insignificant to reward improvement and the denoising meets the required threshold, thereby avoiding redundant costs and preventing over-guidance.
We first denote the step $t$ reward $\mathcal{R}^{(i^*)}_t(\cdot)$ as $r_t$. Then, the smoothed trend of reward gain is defined as:
\begin{equation}
\widetilde{\Delta r}_t=(1-\phi)\,\widetilde{\Delta r}_{t+1}+\phi\,(r_t-r_{t+1}),\quad 
\end{equation}
where $\phi\in(0,1]$.
If and only if $p_t =1$ and $\widetilde{\Delta r}_t \le \delta_r$, the reward return margin has been reached and the early stopping is triggered, where $\delta_r\ge 0$ is the minimal tolerance gain.
We define the early-stopping indicator function as:
\begin{equation}
    \mathrm{E}_{stop}(t)=\mathbb{I}\!\left[p_t =1 \ \wedge\  \widetilde{\Delta r}_t \le \delta_r\right]\in\{0,1\}.
\end{equation}
Exploration and guidance stop when $E_{stop}(t)=1$ and continue when $E_{stop}(t)=0$.
This criterion ensures triggers early stopping only when both conditions are met: $p_t$ guarantees the state is `sufficiently clean', while $\widetilde{\Delta r}_t$ ensures that `further exploration guidance yields no significant reward improvement'. By requiring both conditions simultaneously, this technique eliminates redundant guidance in the later stages of generation without compromising target alignment quality, thereby reducing computational cost.

\section{Experiments}
\subsection{Experimental Setup}
\textbf{Datasets.}
We employ Pick-a-pic~\cite{kirstain2023pick}, HPSv2~\cite{hpsv2}, GenEval~\cite{ghosh2023geneval}, and animals~\cite{black2023training} datasets as our basic test bed. The prompts in Pick-a-Pic are relatively complex, abstract, and surreal, whereas those in HPSv2 are more realistic and
aligned with the real world. GenEval includes specified counting and composition tasks.\\
\textbf{Optimization Objectives and Metrics.} We deploy three reward functions as the optimization objectives, that is, PickScore~\cite{kirstain2023pick} (PS), Aesthetic Score~\cite{aesthetic} (AES), and ImageReward~\cite{xu2023imagereward} (IR). To quantitatively evaluate the generative performance as extensively as possible, we introduce 4 evaluating dimensions with 13 distinct metrics, including:
    {\textit{Aesthetic Preference}}: AES, PS, IR, and HPSv2~\cite{hpsv2}.
    {\textit{Image Fidelity}}: ClipScore~\cite{clip} (Clip), Fréchet Inception Distance (FID)~\cite{fid}, and improved F1 Score (iFS)~\cite{iPR}.
    {\textit{Generative Diversity}}: LPIPS~\cite{LPIPS}, TCE~\cite{TCE}, and Inception Score~\cite{inception} (IS).
    {\textit{Compositional Richness}}: NIQE~\cite{niqe}, BRISQUE (BRI)~\cite{brisque}, and Spectral Entropy (SE)~\cite{spectralE}.
Details of each metric can be found in Appendix.~\ref{metric_appendix}. \\
\textbf{Baselines.} For comprehensive comparison, we introduce three types of State-of-The-Art (SoTA) diffusion-based alignment methods, including:
{\textit{Gradient-based Training}}: Diff-DPO~\cite{wallace2024diffusion}, Diff-KTO~\cite{li2024aligning}, SPO~\cite{liang2024step}, and DRaFT~\cite{DRaFT}.
{\textit{RL-based Training}}: DDPO~\cite{black2023training}, DPOK~\cite{fan2023dpok}.
{\textit{Training Free}}: DAS~\cite{zhang2026inference}, TTScale~\cite{ttscale}, FKSteer~\cite{steer}, DNO~\cite{tang2024inference}, and DyMO~\cite{xie2025dymo}. All baselines are strictly reproduced based on their official code and settings within our evaluation benchmark.\\
\textbf{Backbones.} We adopt two types of widely-adopted backbones: {\textit{Unet-based}} models: SDv14, SDv15, SDv21, SD XL1.0 (XL) and {\textit{Flow-matching}} models: SD 3.0-medium (SD3), SD3.5-medium (SD3.5).

\begin{figure}[b]
    \centering
    \includegraphics[width=\linewidth]{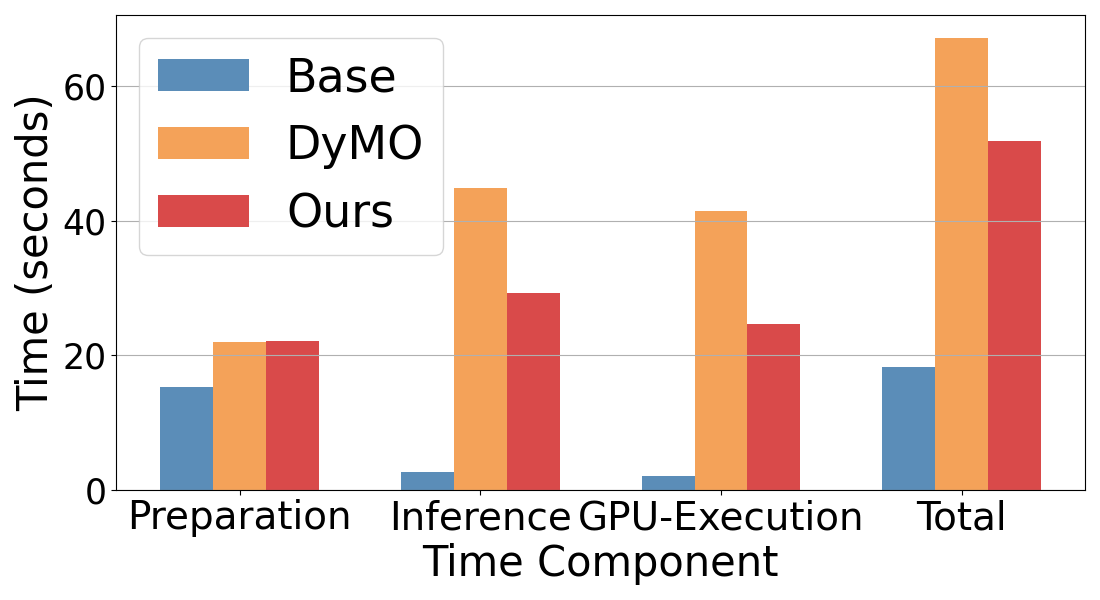}
    \caption{\textbf{Wall-clock runtime comparison} under consistent experimental conditions.}
    \label{fig:wallclock}
\end{figure}
\begin{table}[t]
\centering
\setlength{\tabcolsep}{3.4pt}
\renewcommand{\arraystretch}{1.24}
\caption{Results on HPSv2 dataset.}\label{tab:hpsv2}
\begin{tabular}{lcccccc}\toprule
Method    & IR  & PS & CLIP & LPIPS & TCE & NIQE$\downarrow$ \\\midrule
SDv15     & 0.231 & 21.57 & \textbf{0.249} & 0.643 & 39.62 & 4.656\\\midrule
Diff-DPO  & 0.571 & 22.03 & 0.245 & 0.619 & 37.97 & 4.471\\
SPO       & 0.300 & 21.58 & \textbf{0.249} & 0.595 & 39.37 & 4.172\\
DDPO      & 0.678 & 22.26 & 0.245 & 0.607 & 38.96 & 3.973\\\midrule
DNO       & 0.446 & 21.79 & 0.246 & 0.575 & 39.14 & 4.523\\
DAS       & 0.684 & 23.09 & 0.246 & 0.653 & 39.31 & 4.133\\
DyMO      & 0.805 & 23.46 & 0.246 & 0.610 & 39.57 & 4.088\\\midrule
\rowcolor{blue!15}Ours & \textbf{0.872} & \textbf{23.79} & 0.247 & \textbf{0.671} & \textbf{39.92} & \textbf{2.595}\\ \bottomrule
\end{tabular}
\end{table}

\begin{table}[t]
\caption{Comparison results with advanced \textbf{\textit{flow matching models}}. Gdt represents the baseline that directly applies the reward gradient to the latent code during inference. Ours is designed based on Gdt with the proposed plugin components.}
\label{tab:sd3}
\centering
\small
\setlength{\tabcolsep}{1.2pt}
\renewcommand{\arraystretch}{1.1}
\begin{tabular}{@{}lcccccccccc@{}}
\toprule
\multirow{2}{*}{\textbf{Method}} & \multicolumn{5}{c}{\textbf{SD3}} & \multicolumn{5}{c}{\textbf{SD3.5}} \\ \cmidrule(lr){2-6} \cmidrule(lr){7-11}
 & PS & CLIP & LPIPS & NIQE$\downarrow$ &Top1& PS & CLIP & LPIPS & NIQE$\downarrow$&Top1 \\ \midrule
Base & 21.62 & 0.238 & \textbf{0.645} & 4.312 &1&21.90 &\textbf{0.238} &0.621 &4.553 &1\\
+Gdt & 22.17 & 0.239 & 0.636 & 4.304 &0 &22.95&0.235& 0.616&4.174&0\\ 
\rowcolor{blue!15}+\underline{Ours} & \textbf{22.57} & \textbf{0.241} & 0.614 & \textbf{2.990} &\textcolor{red}{3} &\textbf{23.93} &\textbf{0.238} &\textbf{0.625}&\textbf{3.794}&\textcolor{red}{4}\\
\midrule
\end{tabular}
\end{table}

\begin{figure}[t]
    \centering
    \includegraphics[width=\linewidth]{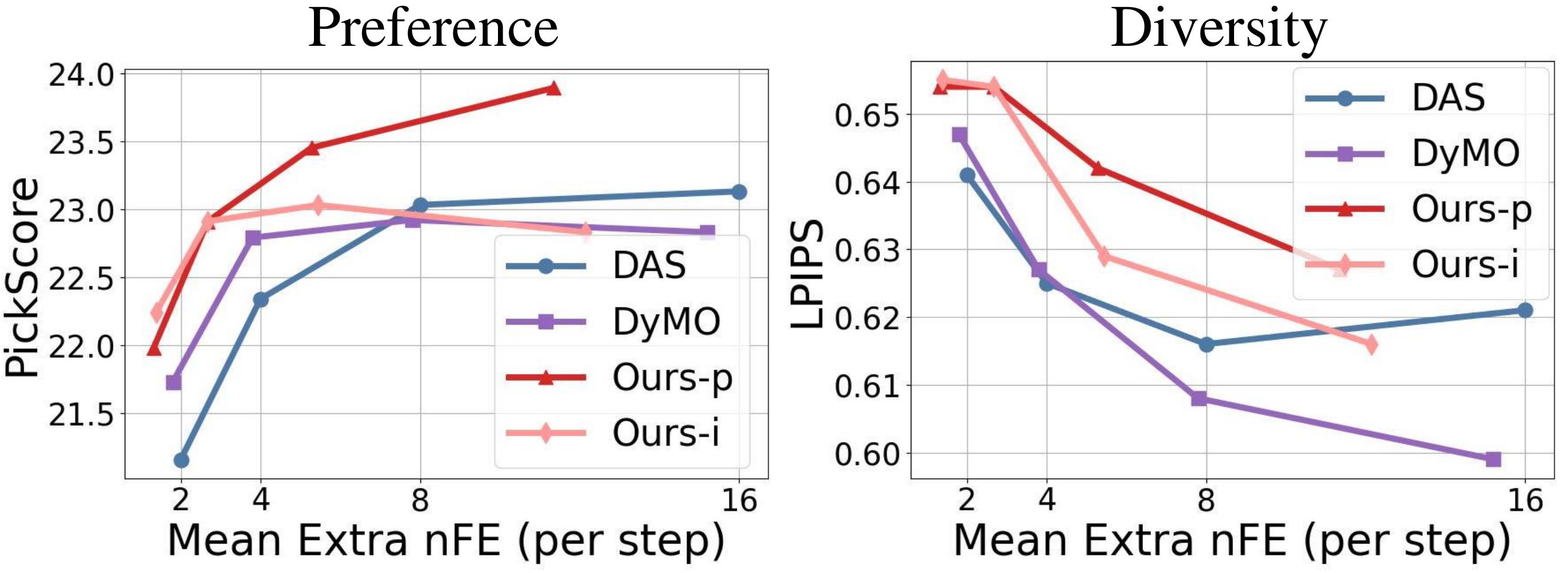}
    \caption{\textbf{Performance-Computation} Trade-off.}
    \label{fig:nFe}
\end{figure}
\begin{figure}[b]
    \centering
    \includegraphics[width=\linewidth]{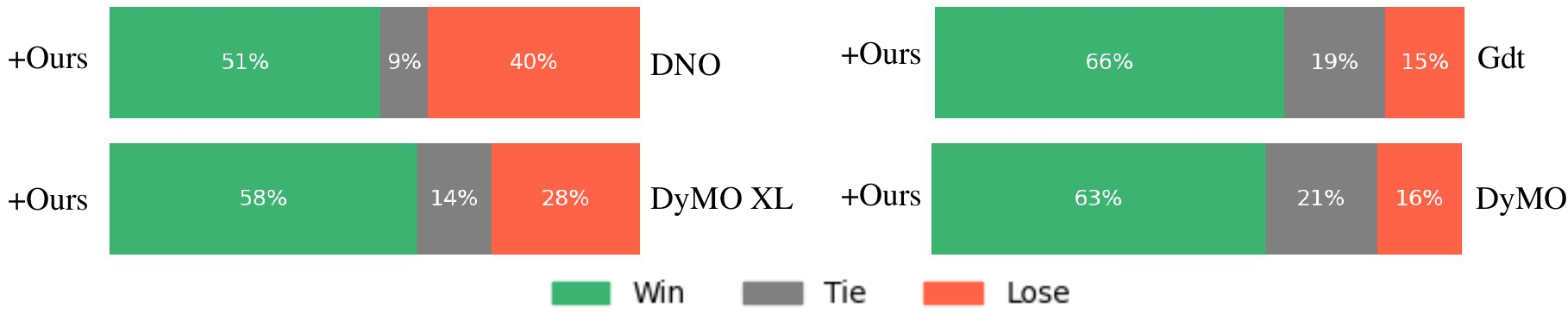}
    \caption{\textbf{User study} for plug-and-play effectiveness.}
    \label{fig:user}
\end{figure}
\begin{figure*}[t]
    \centering
    \includegraphics[width=\linewidth]{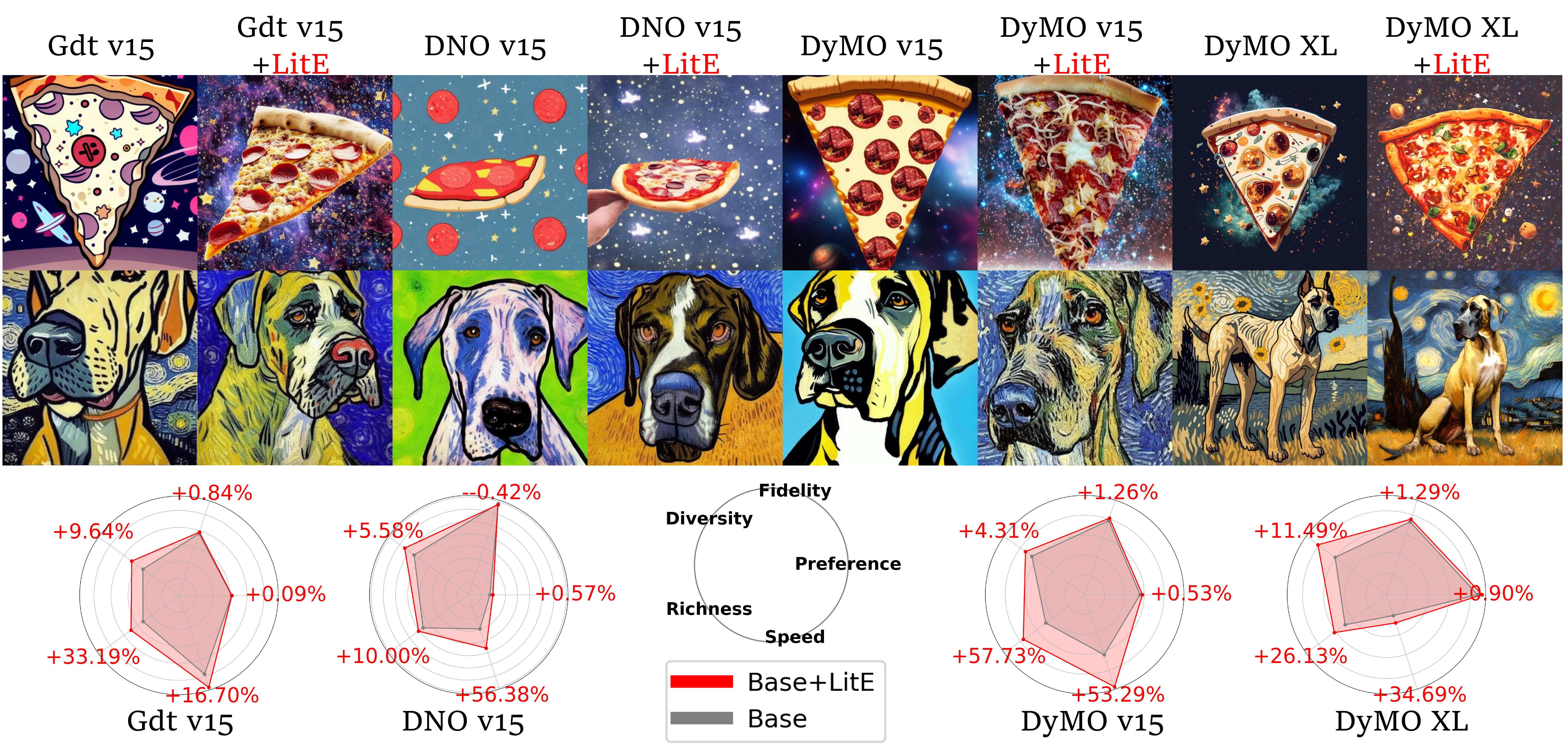}
    \caption{\textbf{Plug-and-Play effectiveness}. While maintaining the Fidelity and Preference, our method significantly enhances Diversity, Richness, and Speed across all cases, such as the ``background and pizza filling diversity'' in the first row, and ``improved Van Gogh style'' in the second row. }
    \label{fig:plugin}
\end{figure*}
\begin{table*}[t!]

    \setlength{\tabcolsep}{6pt} 
\centering
\caption{
\textbf{Ablation study}. For each component, ESV and M-ESV represent exploration supplement variable and with Monte Carlo sampling. I\&R and EarS denote Inheritance-Restart technique and Early Stopping. GuiS denotes Guidance Screening. }\label{tab:abl}
\centering
\begin{tabular}{l|cccccc|cccccccc}
\toprule
 \multirow{2}{*}{Var}  & \multicolumn{2}{c}{Dive} & Fide & \multicolumn{2}{c}{Effi} & Qual         & \multirow{2}{*}{IR} & \multirow{2}{*}{PS} & \multirow{2}{*}{CLIP} & \multirow{2}{*}{LPIPS} & \multirow{2}{*}{TCE} & \multirow{2}{*}{NIQE$\downarrow$} & \multirow{2}{*}{Time$\downarrow$} & \multirow{2}{*}{\#Top2} \\ \cmidrule(lr){2-3} \cmidrule(lr){7-7} \cmidrule(lr){5-6} \cmidrule(lr){4-4}
 & ESV    & M-ESV     & L2     & I\&R& EarS& GuiS &                     &                     &                       &                        &                      &                       &                       &                        \\ \midrule
Base   &          &            &       &          &            &                 &                     0.709&                     22.79&                       0.239&                        0.627&                      39.29&                       4.454&                       44.92&                        0\\ \midrule
V1         & \checkmark       &            &       &          &            &                 &                     0.739&                     22.12&                       0.236&                        \textbf{0.659}&                      39.30&                       2.897&                       46.31&                        1\\
V2         & \checkmark       &            &       \checkmark       
& &            &                 &                     0.734&                     22.46&                       0.240&                        0.636&                      39.30&                       3.379&                       51.24&                        0\\
V3         &          & \checkmark         &       \checkmark       & &            &                 &                     0.757&                     22.75&                       0.240&                        0.641&                      39.32&                       2.939&                       67.08&                        0\\
V4         &          & \checkmark         & \checkmark    & \checkmark       &            &                 &                     0.746&                     22.73&                       0.241&                        0.640&                      39.29&                       3.039&                       48.56&                        0\\
V5         &          & \checkmark         & \checkmark    & \checkmark       & \checkmark         &                 &                     0.759&                     22.66&                       \textbf{0.242}&                        0.645&                      39.28&                       3.120&                       \underline{40.91}&                        2\\
V6         &          & \checkmark         & \checkmark    & \checkmark       &            & \checkmark              &                     \underline{0.774}&                     \textbf{22.95}&                       0.241&                        0.653&                      \underline{39.35}&                       \textbf{2.847}&                       44.73&                        4\\ \midrule
 \rowcolor{blue!15}Ours&          & \checkmark         & \checkmark    & \checkmark       & \checkmark         & \checkmark              &                     \textbf{0.812}&                     \underline{22.91}&                       \textbf{0.242}&                        \underline{0.654}&                      \textbf{39.37}&                       \underline{2.886}&                       \textbf{37.28}&       7\\ \bottomrule                
\end{tabular}

\end{table*}

\begin{figure*}[t!]
    \centering
    \includegraphics[width=\linewidth]{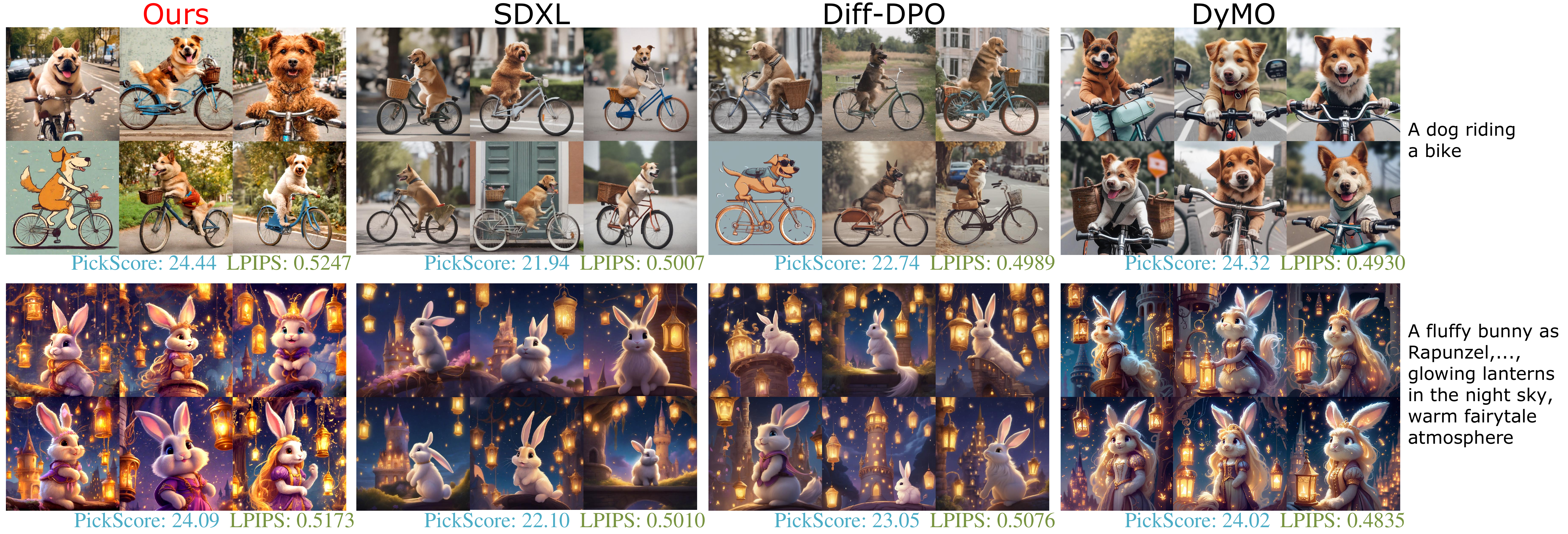}
    \caption{\textbf{Generative diversity examples} with the same prompt and different seeds.}
    \label{fig:diversity}
    \vspace{-0.3cm}
\end{figure*}
\begin{figure*}[t]
    \centering
    \includegraphics[width=\linewidth]{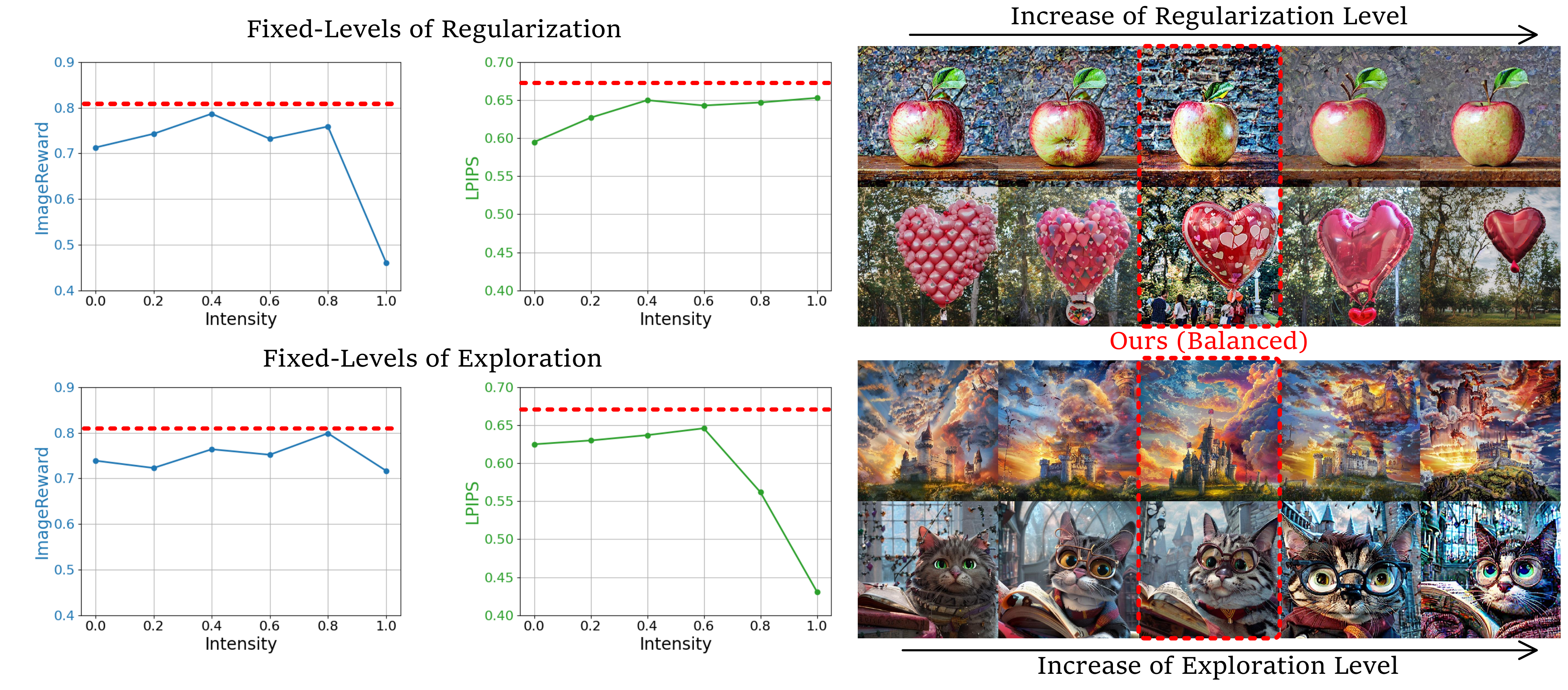}
    \caption{\textbf{Diversity-fidelity adaptive trade-off} (Ours, red dotted lines) with different fixed intensities.}
    \label{fig:tradeoff}
\end{figure*}
\subsection{Comparisons Against SoTA Methods}
To demonstrate the effectiveness of \ourmethod{} (SwiftE), we carefully reproduce and evaluate \textbf{12 baselines} with PickScore as the objective on SDv15 and pick-a-pic datasets. Since SwiftE is an efficient plugin, this section represents deploying SwiftE to DyMO. Then, we introduce \textbf{13 metrics} to assess from four dimensions. As shown in Tab.\ref{tab:main}, our method achieves the best overall performance in all dimensions, \underline{\textit{achieving Top2 performance in 12 metrics}}. Moreover, since the prompts in the Pick-a-pic dataset are relatively abstract and surreal, we further conduct an experiment on the photorealistic HPSv2 dataset for comprehensive evaluation. In Tab.\ref{tab:hpsv2}, all methods perform better because realistic objects are more likely to be seen during training and hence easier to generate. \textit{our method still achieves the best results on five metrics}\textit{and maintains comparable fidelity}. The qualitative results in Appendix Fig.~\ref{fig:main} further demonstrate the superiority of our method. In Appendix.~\ref{seed_appendix} \& Tab.~\ref{tab:metrics_comparison}, to assess the impact of random seeds on result stability, we compute the coefficient of variation (CoV) of each metric for the training-free baseline across different seeds. \\
\noindent{\textbf{Comparisons with more advanced SD models.}}  In Appendix Tab.~\ref{tab:sdxl}, we employ our method and other SoTA on the SD-XL backbone, and introduce the advanced SDv35 as an additional baseline. We also implement it on the advanced SD3 and SD3.5 backbones with PS as the reward and Pick-a-pic as the dataset. Quantitatively, as shown in Tab.~\ref{tab:sd3}, our method can still effectively improve the reward score while maintaining the alignment and diversity. Qualitatively, in Fig.~\ref{fig:sd3}, it can be clearly observed that our method has achieved improved human preference in aesthetics.
\subsection{Plug-and-Play Effectiveness}\label{sec:plug}
In Fig.~\ref{fig:plugin}, we discuss the plugin effectiveness of \ourmethod{}. Specifically, we present side-by-side comparisons of results with and without SwiftE. Gdt represents backbones with Gradient guidance. We choose PS, CLIP, LPIPS, $\frac{1}{\text{NIQE}}$, and image-per-minute as the representations of Preference, Fidelity, Diversity, Richness, and Speed. The results clearly show that SwiftE effectively enhances the diversity of baseline methods and reduces computational cost while improving other quality dimensions. 
\subsection{Analysis on Inference Efficiency}
\textbf{Comparisons on Wall-clock Runtime.} Here, we measure and compare the time consumption of all methods under consistent experimental conditions. To investigate the effectiveness in real-world applications, we conduct comparisons on wall-clock runtime with details. As shown in Fig.~\ref{fig:wallclock}, we present the time consumption of running the entire script. Firstly, their preparation time are similar and once prepared, the script can conduct arbitrary times of inference. Then, it can be observed that our inference time is much faster than the baseline DyMO, while Base has the smallest inference time since it includes no optimization or guidance.\\
\textbf{Performance-Computation Trade-off.}
To investigate the relation between computational cost and performance,  we first deploy \textit{extra mean number of Function Evaluated (nFE)} to quantify the computational cost in a unified manner, which represents the number of extra iterations or sampling in one denoising step. Then, we compare our method with iteration-based DyMO and sampling-based DAS, where the maximum iteration and sampling particle number are set to 2,4,8,16. Notably, \ourmethod{} includes both sampling and iteration, we decouple these factors and design two variants of our method, that is, Ours-i (iteration) and Ours-p (particles). In Fig.~\ref{fig:nFe}, our results can first save nFE because we conduct early stopping. Second, the curves of our results consistently surpass the baselines, indicating the consistent superiority of \ourmethod{}.

\subsection{Ablation Study}

\textbf{Overall Ablation.}
\ourmethod{} aims to balance the trade-offs between diversity-fidelity and quality-efficiency, and thus can be divided into four categories based on the objectives. Specifically, Diversity (Dive) includes directly adding exploration and adding Monte Carlo exploration. Fidelity (Fide) refers to L2 constraints for regularization. Efficiency (Effi) includes the I\&R and early stopping. Guidance Screening aims to enhance Quality (Qual). Then, we conducted detailed ablations in Tab.~\ref{tab:abl}, where the proposed diversity-fidelity tradeoff effectively improves the diversity of generated images while maintaining consistency. Meanwhile, the I\&R mechanism and early stopping greatly reduce computational overhead with almost no impact on performance. GuiS improves inference efficiency and performance by removing invalid guidance. In Appendix.~\ref{ablation_appendix}, we further analyze the mutual correlation among different proposed components based on the ablation results. \\
\textbf{Analysis on Generative Diversity.} The generative diversity is compared among SD-XL, Diff-DPO, DyMO, and Ours by generating images with 40 seeds. As shown in Fig.~\ref{fig:diversity}, our method exhibits improved diversity while maintaining the impressive preference score and precise prompt-image alignment. We have also investigated exploration diversity during inference in Appendix~\ref{app:explore}.\\
\textbf{Analysis on Diversity-Fidelity Trade-off.}
In Fig.~\ref{fig:tradeoff}, we examine the Diversity-Fidelity Trade-off by fixing exploration and regularization levels instead of using dynamic $p_t$.
Results show that stronger regularization undermines preference without enhancing diversity, while higher exploration reduces diversity through image distortion. In contrast, our method consistently surpasses baselines and achieves a balanced trade-off between richness and alignment.

\subsection{User Study}
To subjectively evaluate the generative quality of different methods, we conduct a user study based on the images from the Plug-and-Play experiments in Sec.~\ref{sec:plug}. Specifically, we recruit five subjects without prior knowledge to select the preferred image within paired images with the corresponding prompt as the reference. As shown in Fig.~\ref{fig:user}, our method consistently performs as an effective plugin that achieves superior quality compared to the implemented original models. User Study details are provided in Appendix~\ref{user_appendix}.

\section{Conclusion}

In this work, we address the challenges in aligning diffusion models with specific objectives without incurring prohibitive fine-tuning costs. We introduce \ourmethod{}, a plugin that mitigates distribution collapse and enhances efficiency through an Inheritance-Restart exploration mechanism and a Quality--Efficiency arbitration framework. These innovations prevent early convergence, increase exploration for high-reward trajectories, and improve guidance by eliminating irrelevant signals. Experiments show consistent gains in quality, diversity, and efficiency over existing methods.



\clearpage

\bibliographystyle{ACM-Reference-Format}

\bibliography{refer.bib}

\clearpage

\section*{Appendix}
\setcounter{section}{0}
\renewcommand{\theHsection}{appendix.\arabic{section}}
\renewcommand{\theHsubsection}{appendix.\arabic{section}.\arabic{subsection}}

\begin{figure*}[!t]
  \centering
  \includegraphics[width=\linewidth]{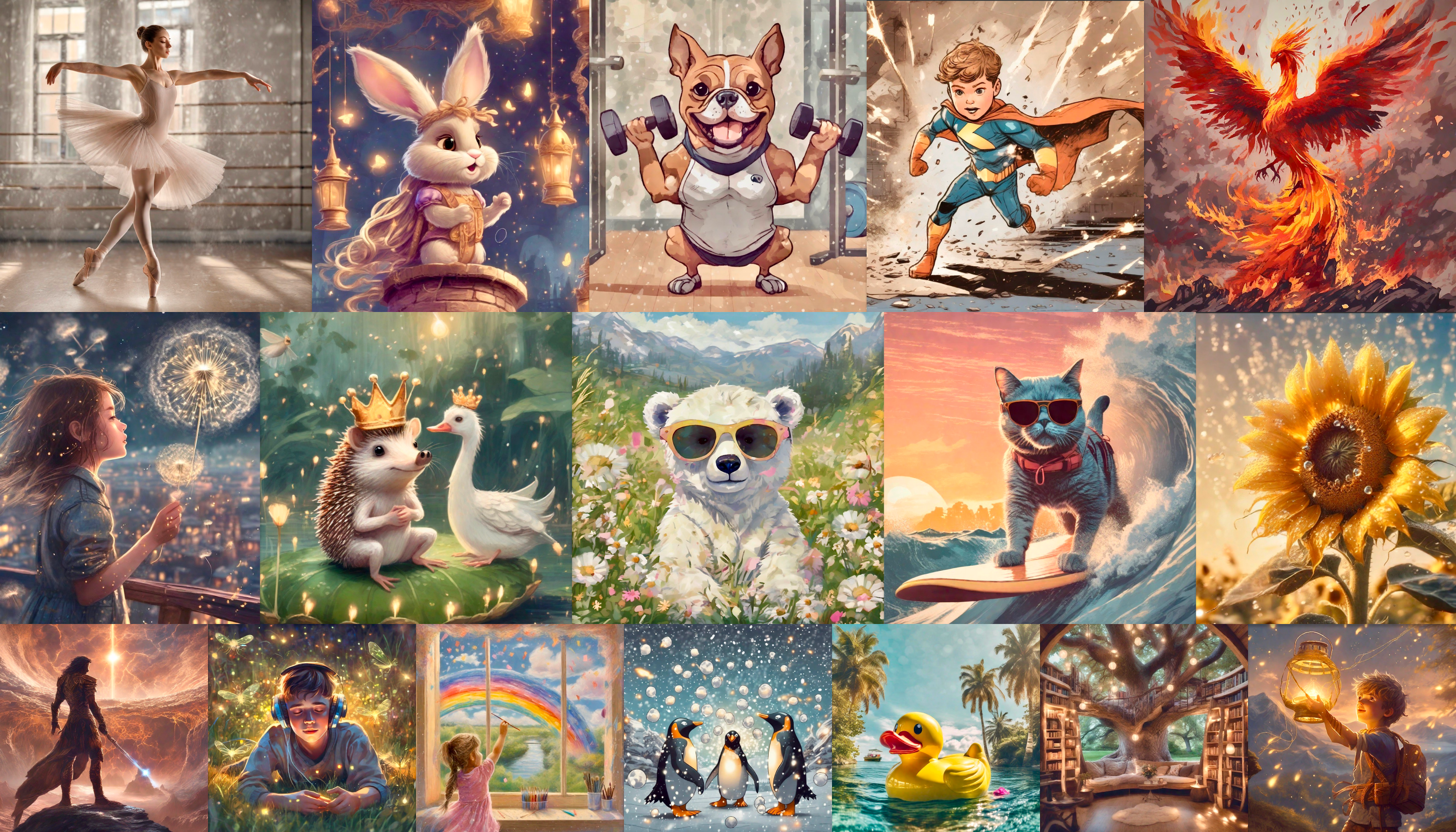}
  \caption{\textbf{Showcase of generated samples.} The images from our model achieve state-of-the-art performance.}
  \label{fig:fig1}
\end{figure*}

\section{Related Work}
\label{related work}

\subsection{Diffusion Model}
Diffusion models have achieved remarkable progress in generative modeling. Their stability and gradual refinement allow them to approximate complex distributions, producing samples with both high fidelity and diversity. They have shown strong generalization across domains, including text-to-image~\cite{li2024aligning,yuan2024self,ding2025rass}, speech~\cite{kong2020diffwave,chen2020wavegrad,liu2022diffsinger}, molecular design~\cite{xu2022geodiff,xu2023geometric}, and protein generation~\cite{watson2023novo,abramson2024accurate,qiao2022dynamic}. However, since these models are typically pretrained on broad datasets lacking task-specific signals, they often fail to reflect user-desired distributions in specialized applications. This gap highlights the need for alignment:enabling customized optimization toward target distributions without losing generality.

\subsection{Gradient-based Fine-tuning Methods}
Direct gradient-based fine-tuning is intuitively feasible for enhancing diffusion models. For examples,
DRaFT~\cite{DRaFT} performs differentiable reward optimization for diffusion models by backpropagating reward gradients through sampling steps, forming a unified gradient-based framework for preference-aligned fine-tuning; DRTune~\cite{drtune} stabilizes reward-supervised diffusion training by applying stop-gradient to denoiser inputs and uniformly sampling K timesteps, enabling efficient supervision across all sampling depths; Diffusion-RPO~\cite{diffrpo} extends relative preference optimization to diffusion models by conducting stepwise preference contrasts with CLIP-based multimodal weighting to learn coherent cross-semantic preference patterns. All these methods improve preference alignment with limited stability, scalability, and robustness under diverse real-world undifferentiated reward signals.

\subsection{Fine-tuning Methods}
Existing alignment approaches fall into two categories: fine-tuning~\cite{black2023training,fan2023dpok} and training-free methods. The mainstream fine-tuning , for example, relies on reinforcement learning (RL)~\cite{sutton2018reinforcement} to adjust diffusion model weights. RL greatly improves alignment to downstream targets but faces challenges: rewards are only available after the full denoising process, making them sparse and prone to reward hacking. This often leads to higher reward scores but reduced diversity and poor distribution coverage. Moreover, RL-based fine-tuning comes with a high computational cost~\cite{kim2025test}.

\subsection{Traing-free Methods}
Training-free methods~\cite{kim2025test,xie2025dymo} provide an alternative: instead of changing diffusion model weights, they bias the sampling process at inference to guide results toward target objectives. For examples, TTScale~\cite{ttscale} boosts generation quality by shifting extra inference compute from longer denoising chains to noise-search guided by a lightweight evaluator.
FKSteer~\cite{steer} enables inference-time controllability by using particle-based weighting and resampling to steer diffusion trajectories without fine-tuning. This avoids costly retraining and reduces computation.
First, prior work lacks an adaptive mechanism to balance diversity and fidelity~\cite{dhariwal2021beat}.
Second, existing training-free methods often overemphasize target guidance, forcing trajectories into narrow regions and reducing coverage of the pretrained distribution. While less prone to severe reward hacking, this effect still harms diversity.
Further, guidance signals are typically applied indiscriminately across all steps~\cite{tang2024inference}, without filtering ineffective or saturated signals. This overuse increases inference cost and amplifies noise, sometimes even causing negative guidance effects.

\section{Method Supplement}
\subsection{Control Network}
\label{pernet}
We assign $y=0$ to samples close to pure noise $\mathbf{x}_T$, and $y=1$ to samples close to the target $\mathbf{x}_0$, pretraining the network via the loss defined as follows.
\begin{equation}
    \mathcal{L}(\theta) = - \mathbb{E}_{(\mathbf{x}_t,y)} \left[ y \log f_\theta(\mathbf{x}_t) + (1-y) \log (1 - f_\theta(\mathbf{x}_t)) \right].
\end{equation}

\subsection{Generalizability of Pre-trained Control Network}
\label{control_network}
Considering training a control network could undermine the claim of ``training-free'' of the proposed \ourmethod{}, it is necessary to demonstrate that the Control Network could also be seen as a pre-trained scorer, which can be pre-trained once and be generalizable for all. Therefore, as shown in Tab.~\ref{tab:control}, we pre-train the control network on the simple animal set and predict steps on other datasets. The results demonstrate that the network could easily generalize to other datasets with similar effective performance, which is caused by the fixed scheduler having already pre-set the noise level for different steps in a coarse manner.

\section{Additional Experiment results}
\subsection{Results with alternative optimization targets.}\label{sec:aes-target}
\begin{figure*}[t]
    \centering
    \includegraphics[width=1\linewidth]{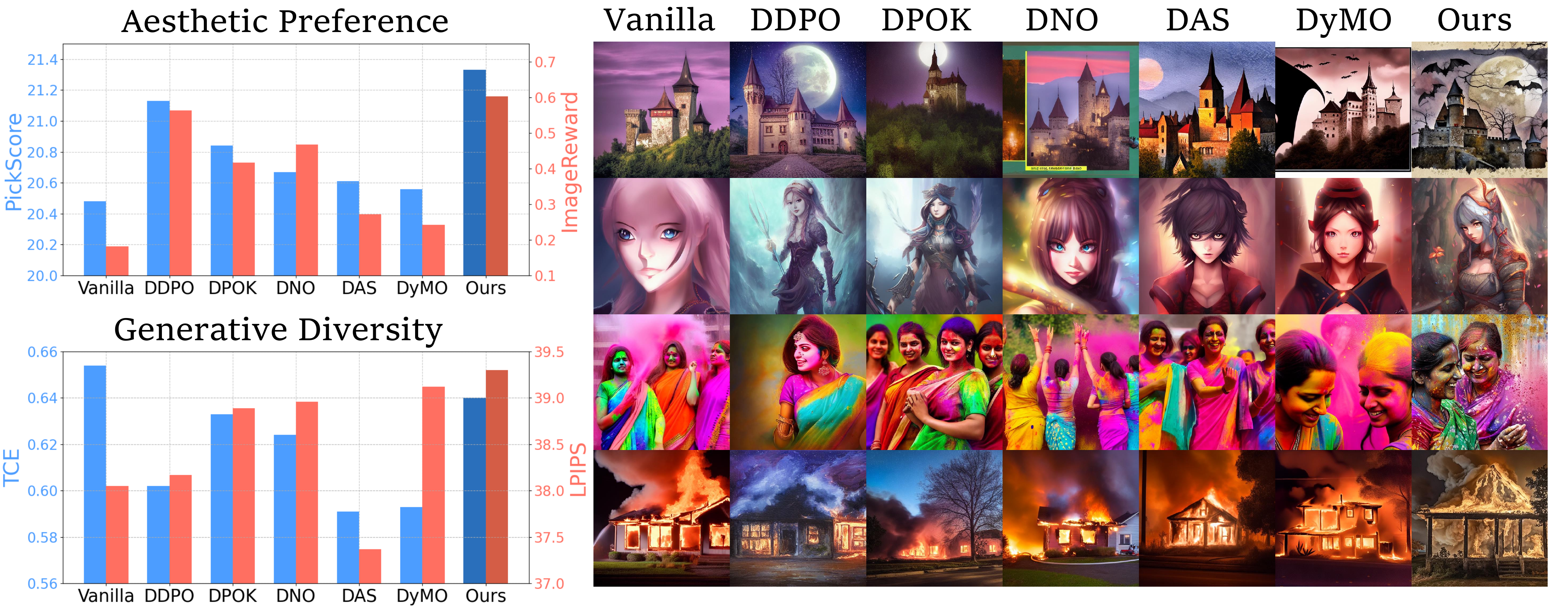}
    \caption{Results with Aesthetic Score as the optimization objective.}
    \label{fig:aes-target}
\end{figure*}

We also conduct experiments by replacing PickScore with Aesthetic score as the optimization target to evaluate the adaptability of \ourmethod{}. As shown in Fig.~\ref{fig:aes-target}, guidance with the aesthetic score demonstrates our overall superiority in aesthetic preference and generative diversity. Then, in the visual results, it could be clearly observed that our generated images contain richer details, better alignment, and improved aesthetic impressions.

\subsection{Detailed hyper-parameter setting}
\begin{table*}[h!]
\centering
\caption{Hyper-parameters and metrics in our experiment. \textbf{\textit{Notably, only $n$ and $\delta_r$ are newly introduced by our method}}; all other hyper-parameters follow DDIM or the baseline methods.
}
\begin{tabular}{@{\hspace{20pt}}l@{\hspace{20pt}}lc@{\hspace{20pt}}}
\toprule
\textbf{Name} & \textbf{Description} & \textbf{Value} \\ \midrule
$\eta$ & eta parameter for the DDIM sampler & 1.0 \\
$w$ & classifier-free guidance weight & 5.0 \\

$m_{dno}$ & Iteration Number for DNO & 20 \\
$m_{dymo}$ & Iteration Number for DyMO & adaptive \\
$n_{das}$ & Particle number for DAS & 4 \\
$mp$ & mixed precision & fp16 \\ \midrule
$n$ (\textbf{\ourmethod-unique}) & Exploration supplement number & 2 \\ 
$\delta_r$ (\textbf{\ourmethod-unique}) & Reward enhancement threshold & 0.1 \\\bottomrule
\end{tabular}
\label{tab:hyper}
\end{table*}

\begin{table}[b]
\centering
\caption{Results of Restart Count and Early Stop across HPSv2 and Pick-a-Pic with SDv15 and SD-XL backbones.}
\label{tab:count}
\begin{tabular}{lccccc}
    \toprule
    \multirow{2}{*}{Avg. Value} & \multirow{2}{*}{Base} & \multicolumn{2}{c}{HPSv2} & \multicolumn{2}{c}{Pick-a-Pic} \\
    \cmidrule(lr){3-4} \cmidrule(lr){5-6}
    & & SDv15 & SD-XL & SDv15 & SD-XL \\
    \midrule
    Restart Count &  50& 4.37 & 8.55 & 7.32 & 10.59 \\
    Guiding Step  &  50& 37.35 & 44.21 & 40.67 & 44.30 \\
    \bottomrule
\end{tabular}
\end{table}
The full list of hyper-parameters in our paper is shown in Table~\ref{tab:hyper}. 

\subsection{Metric}
\label{metric_appendix}
We provide a detailed illustration of each introduced metric:
\begin{itemize}[left=0.5em]
    \item \textbf{Aesthetic Preference}
    \begin{itemize}[leftmargin=2em, labelsep=0.5em]
        \item \textbf{AES (Aesthetic Score)}  
        AES is a metric used for evaluating the aesthetic quality of an image, typically by training a model to predict human aesthetic ratings \cite{aesthetic}. 
        
        \item \textbf{PS (PickScore)}  
        PS is a method based on human selection preferences for rating the aesthetic quality of images \cite{kirstain2023pick}.
        
        \item \textbf{IR (ImageReward)}  
        IR is a metric used for assessing image quality, often in the optimization process of image generation models \cite{xu2023imagereward}.
        
        \item \textbf{HPSv2 (Human Preference Score v2)}  
        HPSv2 is a preference-based metric that predicts human preferences for generated images. This model fine-tunes the CLIP model on the HPD v2 dataset. HPSv2 excels in multiple styles, including animation, concept art, paintings, and photographs \cite{hpsv2}.
    \end{itemize}
    
    \item \textbf{Image Fidelity}
    \begin{itemize}
        \item \textbf{ClipScore}  
        ClipScore is a model-based image scoring metric that evaluates the similarity between a generated image and its textual description. It calculates the cosine similarity between the image and text embeddings. The metric is highly correlated with human judgment and does not require reference text \cite{clip}.
        
        \item \textbf{Fréchet Inception Distance (FID)}  
        FID is a metric used to evaluate the quality of generated images. It compares the mean and covariance of the features extracted from real and generated images using the Inception v3 model. A lower FID value indicates that the generated image is closer to real images \cite{fid}.
        
        \item \textbf{Improved F1 Score (iFS)}  
        iFS is an improvement over the traditional F1 score, aiming to provide a better evaluation of generative model performance. It optimizes the balance between precision and recall for more accurate assessments \cite{iPR}.
    \end{itemize}
    
    \item \textbf{Generative Diversity}
    \begin{itemize}[leftmargin=2em]
        \item \textbf{LPIPS (Learned Perceptual Image Patch Similarity)}  
        LPIPS is a metric for evaluating perceptual similarity between images. It compares the feature activations of images in pre-trained convolutional neural networks (e.g., VGG). A higher LPIPS value indicates lower perceptual similarity and a higher diversity\cite{LPIPS}.
        
        \item \textbf{TCE (Truncated CLIP Entropy)}  
        Truncated CLIP Entropy (TCE) is a measure of semantic diversity of a set of generated images, computed in the joint image–text embedding space of CLIP. It is defined by first mapping each image \( x_i \) to a vector \( \mathbf{z}_i = \mathrm{CLIP}_{\rm img}(x_i) \in \mathbb{R}^d \), then forming the empirical covariance matrix \( \Sigma \) of these vectors, extracting its top-\(k\) eigenvalues \( \lambda_1 \ge \cdots \ge \lambda_k \), and finally computing  
        \[
        \mathrm{TCE}_k = \tfrac{1}{2} \sum_{i=1}^k \log \lambda_i.
        \]  
        TCE gives a tractable proxy for how “spread out” the images are in the CLIP semantic space.
        \cite{TCE}.
        
        \item \textbf{Inception Score (IS)}  
        IS is a metric used to evaluate the quality of generated images. It calculates the entropy of the class distribution of images in a pre-trained Inception v3 model. Higher IS values indicate that the generated images are both clear and diverse \cite{inception}.
    \end{itemize}
    
    \item \textbf{Compositional Richness}
    \begin{itemize}[leftmargin=2em, labelsep=0.5em]
        \item \textbf{NIQE (Natural Image Quality Evaluator)}  
        NIQE is a no-reference image quality assessment metric. It evaluates the image quality by measuring the distance between the natural scene statistics of the image and a natural image database of undistorted images \cite{niqe}. A lower NIQE value indicates better image quality.
        
        \item \textbf{BRISQUE (Blind/Referenceless Image Spatial Quality Evaluator)}  
        BRISQUE is another no-reference image quality metric. It evaluates image quality by analyzing spatial-domain features. It does not rely on reference images and is suitable for a variety of image quality assessment tasks \cite{brisque}.
        
        \item \textbf{Spectral Entropy (SE)}  
        SE is a metric for evaluating the spectral properties of images. It calculates the entropy value of an image’s frequency spectrum to assess the complexity and richness of image details. Higher SE values generally indicate more detailed and textured images \cite{spectralE}.
    \end{itemize}
    
\end{itemize}
\begin{figure*}[t!]
    \centering
    \includegraphics[width=1\linewidth]{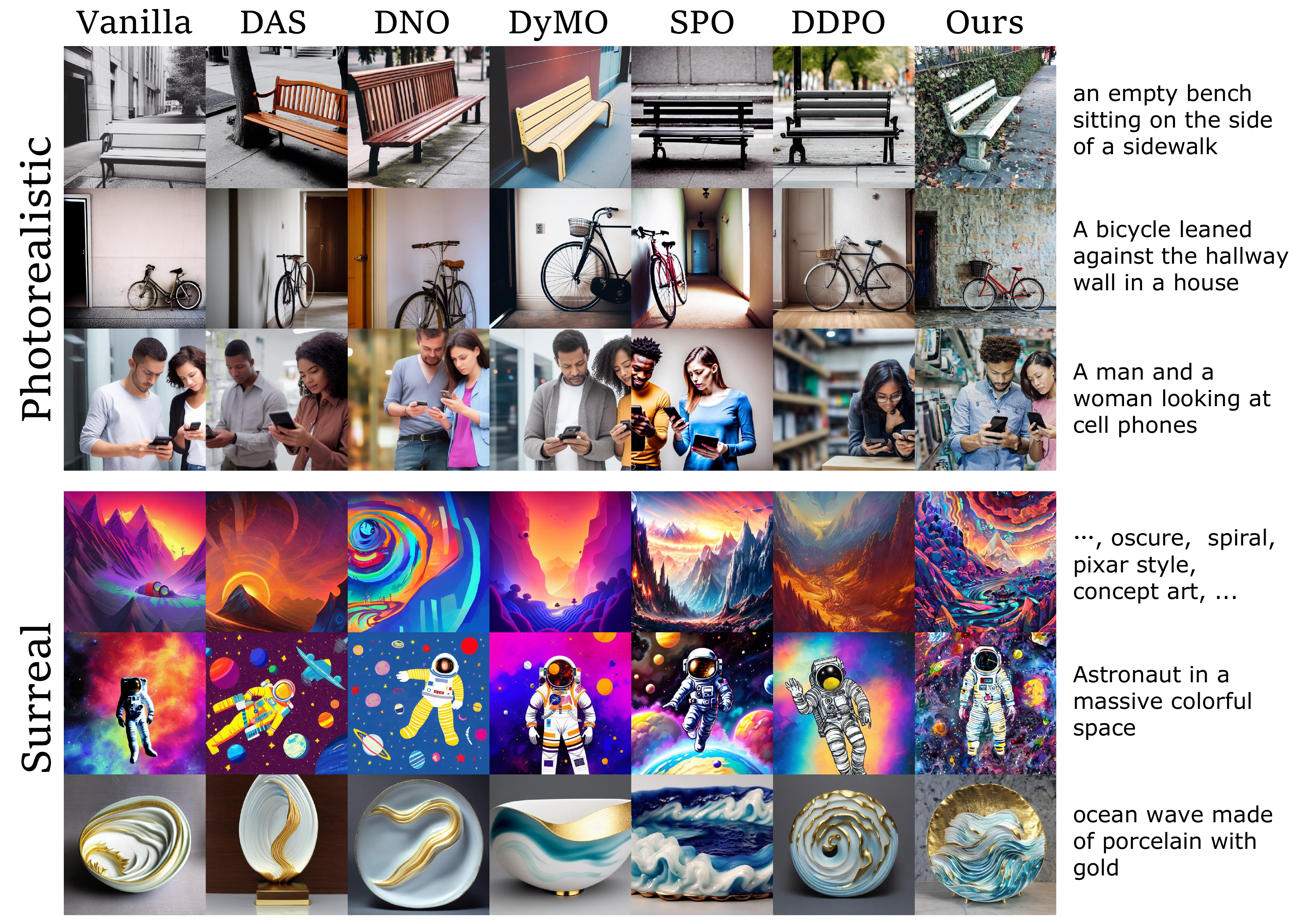}
    \caption{The visual comparisons of different methods on SDv15. For photorealistic prompts from HPSv2, our results contain richer details while maintaining image-prompt alignment. For surreal prompts from Pick-a-Pic, we exhibit much more diverse patterns and colors.}
    \label{fig:main}
\end{figure*}

\begin{table}[htbp]
\centering
\caption{CoV (\%) comparison of methods across four evaluation metrics.}
\label{tab:metrics_comparison}
\begin{tabular}{lcccc}
\toprule
\textbf{Method} & \textbf{Preference} & \textbf{Fidelity} & \textbf{Diversity} & \textbf{Richness} \\
\midrule
DNO  & 0.88 & 0.52 & 0.27 & 1.27 \\
DAS  & 0.96 & 0.81 & 0.30 & 1.36 \\
DyMO & 0.62 & 0.86 & 0.28 & 1.15 \\
Ours & 0.65 & 0.79 & 0.28 & 1.41 \\
\bottomrule
\end{tabular}

\end{table}
\subsection{Supplementary Statistical Significance}
\label{seed_appendix}
Since the generative quality is highly related to the initial noise decided by the seeds, we further conduct experiments with 5 different seeds and calculate their average Coefficient of Variation (CoV). The metrics and evaluation protocol are strictly following Tab.~\ref{tab:main}. As shown in Tab.~\ref{tab:metrics_comparison}, it can be observed that all methods exhibit similarly low CoV across various seeds, which indicates that the guidance-based training-free scaling methods perform relatively stably. These results further demonstrate that the performance reported in Tab. \ref{tab:main} is statistically significant.

\begin{table}[htbp]
\centering
\setlength{\tabcolsep}{2.2pt}
\caption{Mean step numbers predicted by the control network across different datasets. The control network is trained on the simple animal and directly generalizes to other datasets.}
\label{tab:control}
\begin{tabular}{lccccc}
\toprule
\textbf{Method} & \textbf{Total} & \textbf{SimpleAnimal} & \textbf{Pick-a-Pic} & \textbf{HPSv2} & \textbf{GenEval} \\
\midrule
SD15  & 50   & 37.3 & 40.9 & 39.8 & 38.1 \\
SDXL  & 30   & 24.7 & 24.9 & 25.1 & 24.6 \\
\bottomrule
\end{tabular}

\end{table}

\begin{figure}
    \centering
    \includegraphics[width=1\linewidth]{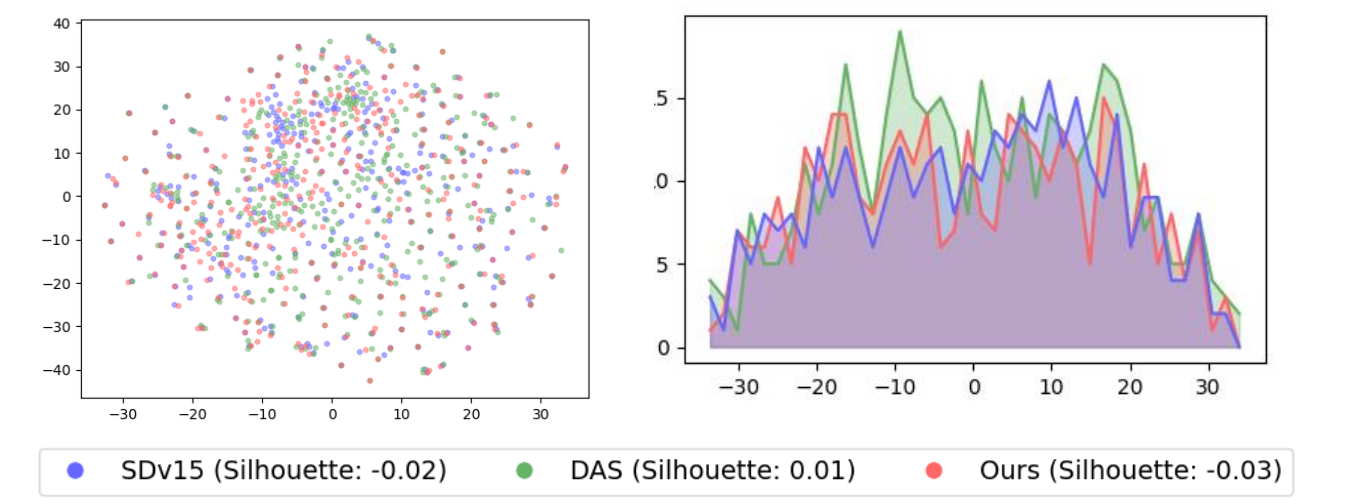}
    \caption{Feature visualization comparison with DAS.}
    \label{fig:das-intro}
\end{figure}
\subsection{Supplementary Distribution Comparison with DAS}
In Fig.~\ref{fig:instr}, we compare the feature distribution with the iteration-based DyMO, thus demonstrating the diversity of our method. Here, we conduct a similar experiment to evaluate the sampling-based baseline, \textit{i.e.}, DAS. As shown in Fig.~\ref{fig:das-intro}, although the sampling-based DAS has richer diversity compared to DyMO, it still tightens the distribution range of the original backbone. In contrast, our result exhibits consistent diversity, which is demonstrated by the wider visualized distribution and the lower Silhouette Coefficient value.

\begin{figure*}[ht]
    \centering
    \includegraphics[width=1\linewidth]{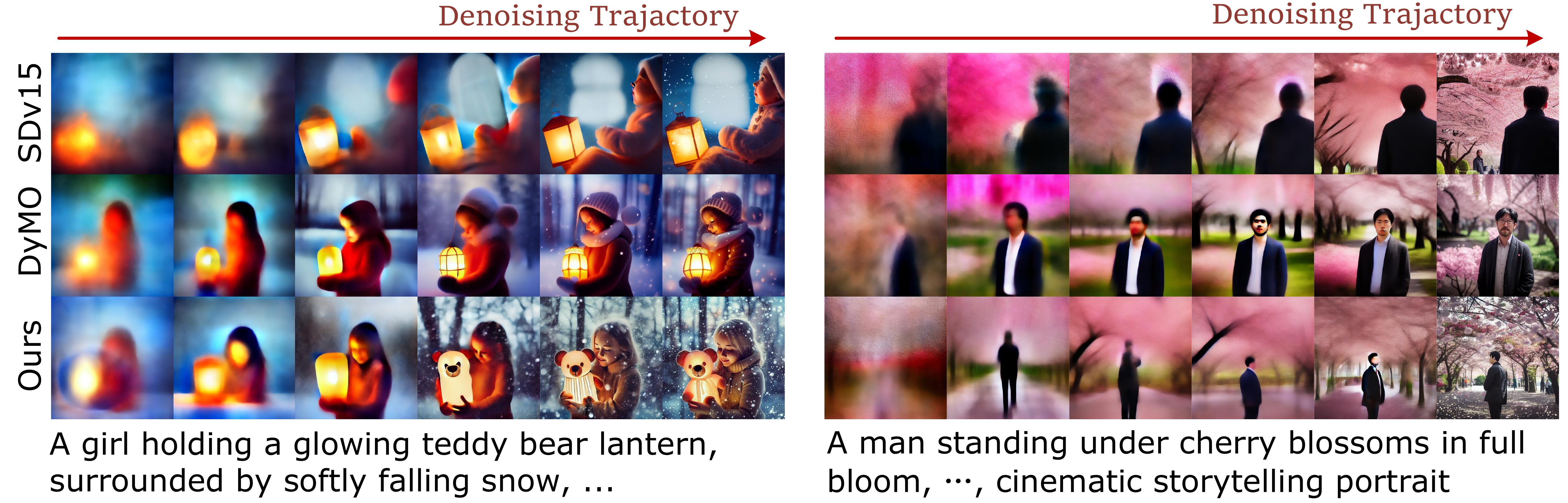}
    \caption{The proposed ESV facilitates exploration during the denoising process.\textbf{ Left}: ESV enables our method to identify a broader range of patterns and adjust the trajectory toward the “teddy bear lantern” during intermediate stages. \textbf{Right}: Our method exhibits greater posture variation throughout denoising. While this variation is not directly responsible for the superior final generation, it reflects an increased level of exploration. }
    \label{fig:Exploration}
\end{figure*}
\subsection{Analysis on Exploration.}\label{app:explore}
To demonstrate the exploration effect of introducing ESV, we present the intermediate results during generation. As shown in Fig.~\ref{fig:Exploration}, our method exhibits superior exploration effect during denoising, implying the improved generative quality. 
\subsection{Further Analysis on Ablation Study}
\label{ablation_appendix}
The proposed \ourmethod{} is designed considering two trade-offs, that is, Diversity-Fidelity and Efficiency-Quality trade-offs. Therefore, different components are introduced with paired correlations. Specifically, introducing M-ESV can introduce exploration and thus improve diversity, while L2 can maintain the fidelity to prevent over-exploration that leads to the collapse of the denoising process. Therefore, without ESV that encourages exploration, solely deploying L2 is unnecessary. Then, both I\&R and EarS are introduced to enhance the generation quality based on the exploration. Meanwhile, GuiS builds upon the previously introduced exploration, as direct guidance typically provides a straightforward path to higher rewards, making GuiS ineffective in such cases. In contrast, with ESV, guidance is not based on the current state but rather incorporates an exploratory shift, allowing GuiS to effectively filter out harmful optimization. As for the intensity of different components in the trade-offs, we present Fig.~\ref{fig:tradeoff} to demonstrate the superiority of our adaptive strategy. Overall, the components of our approach are interdependent and mutually reinforcing, collectively contributing to the superior performance of our image generation scheme in terms of inference-time scaling.

\begin{table*}[t]
\centering
\setlength{\tabcolsep}{3.2pt}
\renewcommand{\arraystretch}{1.06}
\begin{minipage}[t]{0.495\textwidth}
    \centering
    \caption{Results with SDv14.}\label{tab:sd14}
    \begin{tabular}{@{}lcccccc@{}}\toprule
    Method    & IR  & PS & CLIP & LPIPS & TCE & NIQE$\downarrow$ \\\midrule
SDv14     & 0.744 & 20.62 & 0.240 & 0.671 & 40.32 & 4.6312 \\ \midrule
DDPO      & 1.056 & 21.20 & 0.240 & 0.609 & 39.61 & 4.554 \\
DNO       & 0.915 & 21.85 & 0.239 & 0.597 & 39.24 & 4.875 \\
DAS       & 1.217 & 22.58 & 0.238 & 0.610 & 39.56 & 5.451 \\
DyMO      & 1.100 & 23.14 & 0.235 & 0.615 & 39.63 & 4.550 \\ \midrule
\rowcolor{blue!15} Ours     & \textbf{1.230} & \textbf{23.39} & \textbf{0.242} & \textbf{0.676} & \textbf{41.10} & \textbf{3.494} \\
 \bottomrule
    \end{tabular}
\end{minipage}
\hfill
\begin{minipage}[t]{0.495\textwidth}
    \centering
    \caption{Results with SDv21.}\label{tab:sd21}
    \begin{tabular}{@{}lcccccc@{}}\toprule
    Method    & IR  & PS & CLIP & LPIPS & TCE & NIQE$\downarrow$ \\\midrule
SDv21     & 0.541 & 20.63 & 0.280 & 0.644 & 39.69 & 5.672 \\ \midrule
DDPO      & 0.794 & 21.25 & 0.274 & 0.622 & 38.94 & 4.953 \\
DNO       & 0.566 & 20.91 & 0.277 & 0.589 & 39.13 & 5.551 \\
DAS       & 0.817 & 22.29 & 0.280 & 0.656 & 39.55 & 5.150 \\
DyMO      & 0.921 & 22.62 & 0.279 & 0.623 & 39.62 & 5.176 \\ \midrule
\rowcolor{blue!15} Ours     & \textbf{0.980} & \textbf{22.91} & \textbf{0.280} & \textbf{0.677} & \textbf{40.07} & \textbf{3.695} \\ \bottomrule

    \end{tabular}
\end{minipage}
\end{table*}
\subsection{Results on more SD backbones.}
In addition to SDv15 and SD-XL presented in the main paper, we provide the results on SDv14 and SDv21-turbo. As shown in Tab.~\ref{tab:sd14} and Tab.~\ref{tab:sd21}, our method also leads in all metrics.
To further validate robustness on larger and stronger backbones, we additionally report results on advanced models in Tab.~\ref{tab:sdxl}. The same trend remains clear: our method consistently improves preference-related quality while preserving fidelity and diversity. Compared with optimization-heavy baselines, it achieves a better balance between visual richness and prompt alignment, indicating that the plugin design remains effective when scaling to stronger pretrained models.

\begin{table}[t]
\centering
\small
\setlength{\tabcolsep}{3.4pt}
\renewcommand{\arraystretch}{1.14}
\caption{Results with advanced models.}\label{tab:sdxl}
\begin{tabular}{lcccccc}\toprule
Method    & IR  & PS & CLIP & LPIPS & TCE & NIQE$\downarrow$ \\\midrule
SD-XL     & 0.715 & 21.45 & 0.236 & 0.675 & 40.18 & 4.553\\ \midrule
Diff-DPO  & 1.010 & 22.05 & 0.236 & 0.651 & 39.45 & 4.440\\
SPO       & 1.179 & 22.81 & 0.232 & 0.556 & 40.42 & 3.900\\
SDv35     & 1.193 & 21.90 & \textbf{0.238} & 0.621 & 39.42 & 4.873\\\midrule
DNO       & 0.924 & 22.58 & \textbf{0.238} & 0.581 & 39.34 & 4.783\\
DAS       & 1.171 & 23.07 & 0.237 & 0.594 & 39.36 & 5.365\\
DyMO      & 1.079 & 24.34 & 0.233 & 0.609 & 39.55 & 4.510\\\midrule
\rowcolor{blue!15}Ours & \textbf{1.201} & \textbf{24.56} & 0.236 & \textbf{0.679} & \textbf{40.99} & \textbf{3.573}\\ \bottomrule
\end{tabular}
\end{table}
\subsection{Exhibited adaptive values for computation reduction.}
Here, we provide the exhibited average values of restart counts and early stop positions. Specifically, we calculate the average restart counts and average number of guiding steps on HPSv2 and Pick-a-Pic datasets with SDv15 and SD-XL. As shown in Tab.~\ref{tab:count}, it can be observed that HPSv2 has smaller values of both variables compared to Pick-a-Pic, while SDv15 has smaller values than SD-XL. This is because HPSv2 has more photorealistic images that are easier than surreal one for denoising, while SDv15's lesser robustness makes it more amenable to external guidance.
\begin{algorithm}[t]
\caption{Algorithm of Swift Diversity Exploration (\ourmethod{})}\label{alg:main}
		\KwIn{
        Total Inference Denoising Step $T)$;
        Initialized Latent: $\mathbf{x}_{T}$;
        Pre-trained score function $\mathbf{s_{\theta}}(\cdot)$.
        }
        Initialize Inheritance Flag $\text{f}_{inh}=\text{False}$;
        
        \For{ timestep $t \in \{T,...,1\}$}{

            execute one denoising step

$                \mathbf{x}_{t} = \frac{1}{\sqrt{\alpha_{t+1}}} \left( {{\mathbf{x}}}^{*}_{t+1} + \beta_{t+1} \mathbf{s}_\theta({\mathbf{x}}^{*}_{t+1}, {t+1}) \right) + \sqrt{\beta_{t+1}} \mathbf{z}_{t+1}.$
        
          considering early-stopping

            $
            \mathrm{E}_{stop}(t)=\mathbb{I}\!\left[p_t =1 \ \wedge\  \widetilde{\Delta r}_t \le \delta_r\right]\in\{0,1\}.
            $
        
            \If{$\mathrm{E}_{stop}(t)==0$}{
        
            inheritance-restart mechanism for exploration variable

            \If{$\text{f}_{inh}$}{

            inherit previous exploration variable
            
               $\boldsymbol{\epsilon}_{t}^{(i)} = \mathbf{x}_{t+1}^{\star} - {\mathbf{x}}_{t+1}, \quad i=1$
                
            }
            \Else{
                sample new Monte Carlo exploration set $\{\boldsymbol{\epsilon}^{(i)}_t\}$

                $\boldsymbol{\epsilon}^{(i)}_t \sim \mathcal{N}(0, \mathbf{\sigma}_t^2 I), \quad i=1,\dots,n$

            }
            generate candidates with Diversity-Fidelity trade-off
            
            $\hat{\mathbf{x}}_t^{(i)} = {\mathbf{x}}_t + \underbrace{(1-p_t) \cdot \boldsymbol{\epsilon}^{(i)} - p_t\cdot \nabla_{\mathbf{x}_t} L_2(\hat{\mathbf{x}}_t,{\mathbf{x}}_t)}_{\text{Diversity-Fidelity Trade-off}} + \nabla_{\mathbf{x}_t} \mathcal{R}({\mathbf{x}}_{0|t}),$

            iteratively optimizing candidates
            
            $\tilde{\mathbf{x}}_t^{(i)}=g(\hat{\mathbf{x}}_t^{(i)},\mathcal{R}-L_2,m)$

            obtain the optimal candidates
            
            $ \quad \mathbf{x}_t^\star = \tilde{\mathbf{x}}_t^{(i^\star)}, \quad i^\star = \arg\max_{i \in \{1,\dots,n\}} \mathcal{R}(\tilde{\mathbf{x}}_{0|t}^{(i)}).$

            verify optimization condition

            $\Delta \mathcal{R}_t^{(i^{\star})} = \mathcal{R}({\mathbf{x}}_{0|t}^{\star}) - \mathcal{R}({\mathbf{x}}_{0|t}),$

            \If{$\Delta \mathcal{R}_t^{(i^{\star})}>0$}{
                $\text{f}_{inh}=\text{True}$
            }
            \Else{
                $\mathbf{x}^*_t=\mathbf{x}_t$
                
                $\text{f}_{inh}=\text{False}$
            }

            }

}
        
		\KwOut{Optimized Inference Result $\mathbf{x}_0$.}  
\end{algorithm}
\subsection{User Study Details.}
\label{user_appendix}
In this study, we assessed the plug-and-play effectiveness of our method through a human evaluation, where five evaluators compared our method to four competing models: DNO, DyMO XL, Gdt, and DyMO. The evaluation was method-blind, with participants unaware of which model generated each image. The evaluation criteria is:
\begin{itemize}[leftmargin=2em]
    \item \textbf{Win}: Our method was preferred.
    \item \textbf{Tie}: Both methods were considered equally effective, hard to decide distinguishable preference.
    \item \textbf{Lose}: Our method was less preferred.
\end{itemize}

Each participant compared the generated images based on overall quality, with the results shown in Fig.~\ref{fig:user}. The preference for our method across different models demonstrates its superior performance in plug-and-play integration.

\section{Algorithm of the proposed \ourmethod{}}\label{sec:alg}
For improved delivery of the proposed \ourmethod{}, we show its detailed algorithm in Alg.~\ref{alg:main}.

\section{Showcase Prompt Table}
Please refers to Tab.~\ref{tab:labelfig1_part1},~\ref{tab:labelfig1_part2},~\ref{tab:labelfig2},~\ref{tab:labelfig3}, and~\ref{tab:labelfig4}.
\FloatBarrier
\clearpage
\newcolumntype{C}[1]{>{\centering\arraybackslash}m{#1}} 
\newcolumntype{Y}{>{\centering\arraybackslash}m{0.7\linewidth}}

\begin{table}[H]
\centering
\caption{Detailed prompts used for generated images in Fig.~\ref{fig:fig1} - Part 1}\label{tab:labelfig1_part1}
\small
\setlength{\tabcolsep}{6pt}
\renewcommand{\arraystretch}{1.25}
\setlength{\extrarowheight}{2pt}

\begin{tabular}{C{0.22\linewidth}|Y}
\toprule
\textbf{Image} & \textbf{Prompt} \\
\midrule
Fig.~\hyperref[fig:fig1]{1}, Row~1, Col~1 & ballet dancer, insanely detailed, photorealistic, 8k, perfect composition, volumetric lighting, natural complexion, award-winning professional photography, taken with Canon EOS 5D Mark IV, 85mm, mindblowing, masterpiece \\ \hline
Fig.~\hyperref[fig:fig1]{1}, Row~1, Col~2 & A fluffy bunny as Rapunzel, with long golden ears flowing down from a tall enchanted tower, glowing lanterns in the night sky, warm fairytale atmosphere \\ \hline
Fig.~\hyperref[fig:fig1]{1}, Row~1, Col~3 & A dog in sportswear lifting tiny dumbbells at the gym, determined expression, humorous fitness illustration \\ \hline
Fig.~\hyperref[fig:fig1]{1}, Row~1, Col~4 & A boy superhero landing on the ground in classic ``hero pose,'' debris and glowing sparks flying around, comic action shot \\ \hline
Fig.~\hyperref[fig:fig1]{1}, Row~1, Col~5 & A phoenix rising up from ashes \\ \hline
Fig.~\hyperref[fig:fig1]{1}, Row~2, Col~1 & A young girl standing on a rooftop, blowing dandelions that transform into glowing comets, shooting across the night sky, dreamy fantasy artwork \\ \hline
Fig.~\hyperref[fig:fig1]{1}, Row~2, Col~2 & A small hedgehog as the Frog Prince, wearing a tiny crown while sitting on a lilypad, a kind-hearted swan princess leaning close, surrounded by glowing fireflies, magical fairytale illustration \\ \hline
Fig.~\hyperref[fig:fig1]{1}, Row~2, Col~3 & a white polar bear cub wearing sunglasses sits in a meadow with flowers. \\ \hline
Fig.~\hyperref[fig:fig1]{1}, Row~2, Col~4 & A cat surfing on a giant wave at sunset, wearing cool shades, cinematic sports illustration \\ \hline
Fig.~\hyperref[fig:fig1]{1}, Row~2, Col~5 & A sunflower in full bloom under golden sunlight, tiny dewdrops sparkling on its petals, cinematic macro fantasy illustration \\
\bottomrule
\end{tabular}

\end{table}
\begin{table}[b]
\centering
\caption{Detailed prompts used for generated images in Fig.~\ref{fig:tradeoff}}\label{tab:labelfig4}
\small
\setlength{\tabcolsep}{6pt}
\renewcommand{\arraystretch}{1.25}
\setlength{\extrarowheight}{2pt}

\begin{tabular}{C{0.22\linewidth}|Y}
\toprule
\textbf{Image} & \textbf{Prompt} \\
\midrule
Fig.~\ref{fig:tradeoff}, Row~1 & Heart shaped balloon \\ \hline
Fig.~\ref{fig:tradeoff}, Row~2 & An apple on a table \\ \hline
Fig.~\ref{fig:tradeoff}, Row~3 & A castle in the sky, clouds, sunset, explosion \\ \hline
Fig.~\ref{fig:tradeoff}, Row~4 & Harry potter as a cat, pixar style, octane render, HD, high-detail\\
\bottomrule
\end{tabular}

\end{table}

\begin{table}[H]
\centering
\caption{Detailed prompts used for generated images in Fig.~\ref{fig:fig1} - Part 2}\label{tab:labelfig1_part2}
\small
\setlength{\tabcolsep}{6pt}
\renewcommand{\arraystretch}{1.25}
\setlength{\extrarowheight}{2pt}

\begin{tabular}{C{0.22\linewidth}|Y}
\toprule
\textbf{Image} & \textbf{Prompt} \\
\midrule
Fig.~\hyperref[fig:fig1]{1}, Row~3, Col~1 & A warrior standing at the edge of a glowing crater, surrounded by swirling cosmic energy, their silhouette outlined against the birth of a new star, ultimate epic fantasy art \\ \hline
Fig.~\hyperref[fig:fig1]{1}, Row~3, Col~2 & A boy lying on the grass in a field, listening to music with glowing headphones, fireflies surrounding him \\ \hline
Fig.~\hyperref[fig:fig1]{1}, Row~3, Col~3 & A little girl painting a rainbow bridge from the classroom window into the sky, playful magical fairytale art, hopeful and inspiring \\ \hline
Fig.~\hyperref[fig:fig1]{1}, Row~3, Col~4 & A group of playful penguins throwing glowing snowballs at each other, each snowball turning into sparkling stars when it explodes, magical fairytale scene \\ \hline
Fig.~\hyperref[fig:fig1]{1}, Row~3, Col~5 & Giant rubber duck floating in the ocean with a small island on its back, surrounded by tropical palm trees and crystal clear water, bright and sunny day, calm seas, vivid colors, cinematic lighting, high detail \\ \hline
Fig.~\hyperref[fig:fig1]{1}, Row~3, Col~6 & A cozy library built inside an ancient oak tree, warm lights glowing through round windows, whimsical fairytale healing atmosphere \\ \hline
Fig.~\hyperref[fig:fig1]{1}, Row~3, Col~7 & A brave boy carrying a glowing lantern, releasing trails of light that form a golden sunrise, cinematic epic fantasy style \\
\bottomrule
\end{tabular}

\end{table}

\begin{table}[H]
\centering
\caption{Detailed prompts used for generated images in Fig.~\ref{fig:aes-target}}\label{tab:labelfig2}
\small
\setlength{\tabcolsep}{6pt}
\renewcommand{\arraystretch}{1.25}
\setlength{\extrarowheight}{2pt}

\begin{tabular}{C{0.22\linewidth}|Y}
\toprule
\textbf{Image} & \textbf{Prompt} \\
\midrule
Fig.~\ref{fig:aes-target}, Row~1 & transylvania castle on hiltop and dusk bats and scifi \\ \hline
Fig.~\ref{fig:aes-target}, Row~2 & fantasy character portrait digital painting, anime style, detailed with beautiful emotive lighting suggesting personality, and background that suggests character backstory \\ \hline
Fig.~\ref{fig:aes-target}, Row~3 & Women in Saree playing Holi \\ \hline
Fig.~\ref{fig:aes-target}, Row~4 & a house burning at night \\
\bottomrule
\end{tabular}
\end{table}

\begin{table}[H]
\centering
\caption{Detailed prompts used for generated images in Fig.~\ref{fig:plugin}}\label{tab:labelfig3}
\small
\setlength{\tabcolsep}{6pt}
\renewcommand{\arraystretch}{1.25}
\setlength{\extrarowheight}{2pt}

\begin{tabular}{C{0.22\linewidth}|Y}
\toprule
\textbf{Image} & \textbf{Prompt} \\
\midrule
Fig.~\ref{fig:plugin}, Row~1 & a slice of pizza floating through space with stars in the background \\ \hline
Fig.~\ref{fig:plugin}, Row~2 & A Great Dane dog in the style of Vincent Van Gogh \\
\bottomrule
\end{tabular}

\end{table}

\end{CJK*}

\end{document}